\documentclass[10pt,twocolumn,letterpaper]{article}

\usepackage{iccv}      

\usepackage{graphicx}
\usepackage{amsmath}
\usepackage{amssymb}
\usepackage{booktabs}
\usepackage{hyperref}
\hypersetup{breaklinks=true,letterpaper=true,colorlinks}

\iccvfinalcopy 

\def\iccvPaperID{10168} 

\usepackage[skip=4pt]{caption}
\usepackage{listings}
\usepackage{fancyhdr}
\usepackage{multirow}
\usepackage{arydshln}
\usepackage{pifont}
\usepackage{times}
\usepackage{url}            
\usepackage{booktabs}       
\usepackage{amsfonts}       
\usepackage{soul,colortbl}
\usepackage{algpseudocode}
\usepackage{algorithm}
\usepackage{enumitem}
\graphicspath{ {./images/} }

\definecolor{codegreen}{rgb}{0,0.6,0}
\definecolor{codegray}{rgb}{0.5,0.5,0.5}
\definecolor{codepurple}{rgb}{0.58,0,0.82}
\definecolor{backcolour}{rgb}{0.95,0.95,0.92}

\lstdefinestyle{mystyle}{
    backgroundcolor=\color{backcolour},   
    commentstyle=\color{codegreen},
    keywordstyle=\color{magenta},
    numberstyle=\tiny\color{codegray},
    stringstyle=\color{codepurple},
    basicstyle=\ttfamily\scriptsize,
    breakatwhitespace=false,         
    breaklines=true,                 
    captionpos=b,                    
    keepspaces=true,                 
    numbers=left,                    
    numbersep=5pt,                  
    showspaces=false,                
    showstringspaces=false,
    showtabs=false,                  
    tabsize=2
}

\definecolor{tblue}{RGB}{23, 73, 146}
\definecolor{bblue}{RGB}{147, 205, 221}
\definecolor{Gray}{gray}{0.9}
\definecolor{dred}{RGB}{192, 0, 0}

\newcommand{\RN}[1]{%
	\textup{\lowercase\expandafter{\it \romannumeral#1}}%
}
\ificcvfinal\fi

\usepackage[capitalize]{cleveref}
\crefname{section}{Sec.}{Secs.}
\Crefname{section}{Section}{Sections}
\Crefname{table}{Table}{Tables}
\crefname{table}{Tab.}{Tabs.}

\begin{document}

\title{Towards Authentic Face Restoration with Iterative Diffusion Models and Beyond}

\author{}
\author{Yang Zhao\quad Tingbo Hou\quad  Yu-Chuan Su\quad Xuhui Jia\quad Yandong Li \quad  Matthias Grundmann \\
Google\\
{\hspace{-0.3cm}\tt\footnotesize \{yzhaoeric,\,tingbo,\,ycsu,\,xhjia,\,yandongli,\,grundman\}@google.com}
}

\maketitle
\ificcvfinal\thispagestyle{empty}\fi
\begin{abstract}
An authentic face restoration system is becoming increasingly demanding in many computer vision applications, e.g., image enhancement, video communication, and taking portrait. Most of the advanced face restoration models can recover high-quality faces from low-quality ones but usually fail to faithfully generate realistic and high-frequency details that are favored by users. To achieve authentic restoration, we propose \textbf{IDM}, an \textbf{I}teratively learned face restoration system based on denoising \textbf{D}iffusion \textbf{M}odels (DDMs). We define the criterion of an authentic face restoration system, and argue that denoising diffusion models are naturally endowed with this property from two aspects: intrinsic iterative refinement and extrinsic iterative enhancement. Intrinsic learning can preserve the content well and gradually refine the high-quality details, while extrinsic enhancement helps clean the data and improve the restoration task one step further. We demonstrate superior performance on blind face restoration tasks. Beyond restoration, we find the authentically cleaned data by the proposed restoration system is also helpful to image generation tasks in terms of training stabilization and sample quality. Without modifying the models, we achieve better quality than state-of-the-art on FFHQ and ImageNet generation using either GANs or diffusion models.
\end{abstract}

\section{Introduction}
\label{sec:intro}

\begin{figure*}
\centering
\resizebox{0.99\linewidth}{!}{
\begingroup
\setlength{\tabcolsep}{1pt} 
\renewcommand{\arraystretch}{1.0} 
    \begin{tabular}{c|ccc}
        \toprule
        Source & Codeformer~\cite{zhou2022towards} & GFPGAN\cite{wang2021towards} & IDM (Ours) \\ \midrule
        \includegraphics[width=0.24\linewidth]{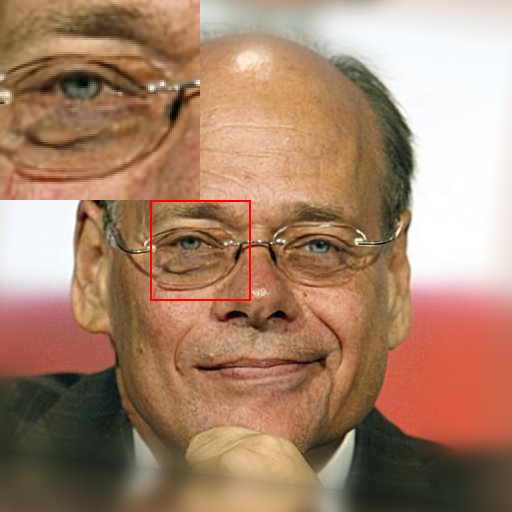} &
        \includegraphics[width=0.24\linewidth]{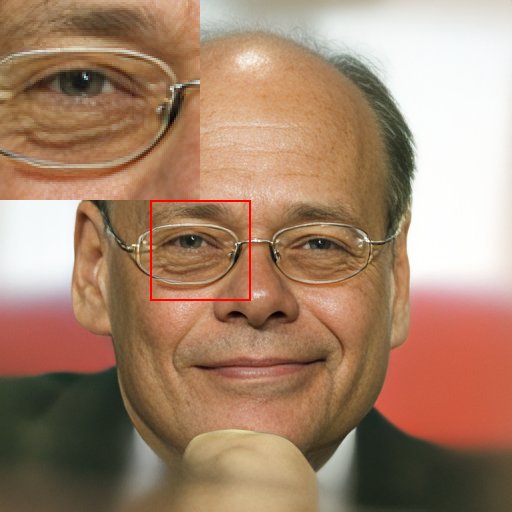} &
        \includegraphics[width=0.24\linewidth]{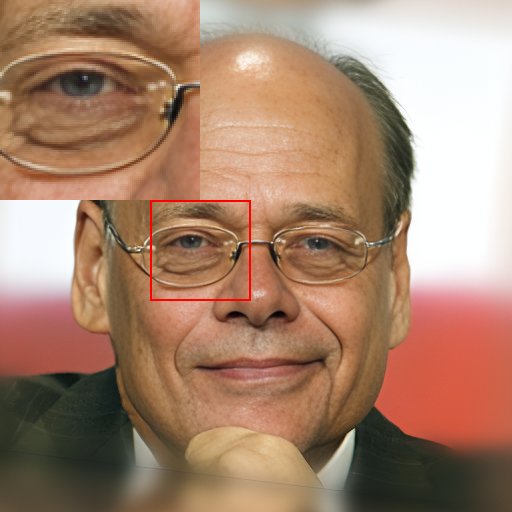} & 
        \includegraphics[width=0.24\linewidth]{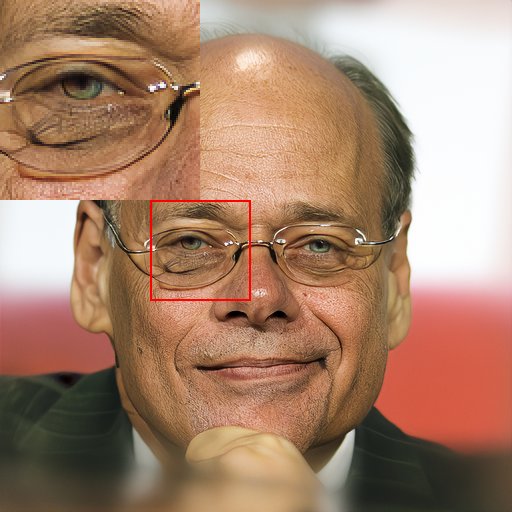}\\
        \includegraphics[width=0.24\linewidth]{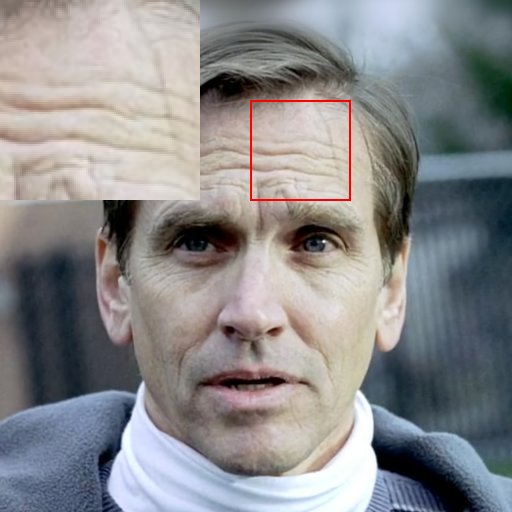} &
        \includegraphics[width=0.24\linewidth]{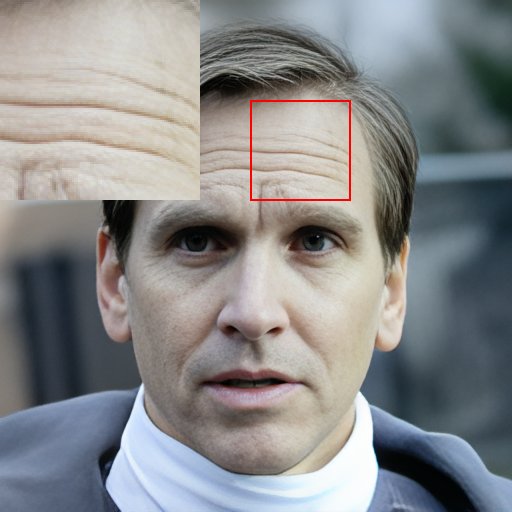} &
        \includegraphics[width=0.24\linewidth]{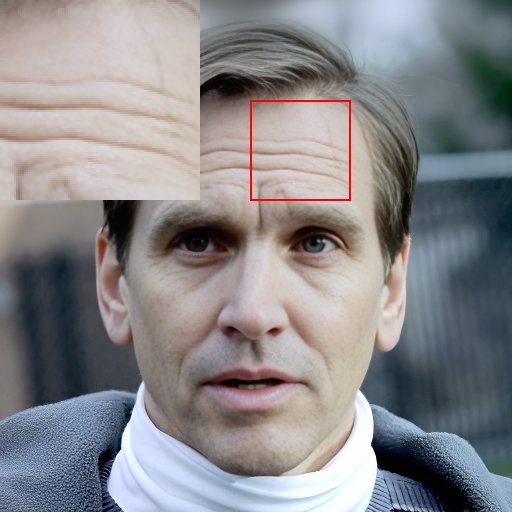} & 
        \includegraphics[width=0.24\linewidth]{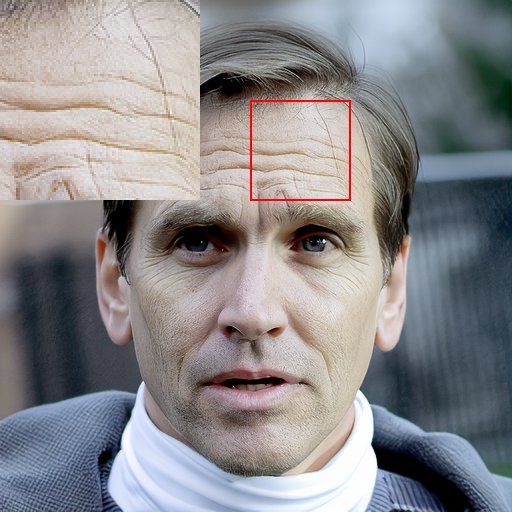}\\         
        \includegraphics[width=0.24\linewidth]{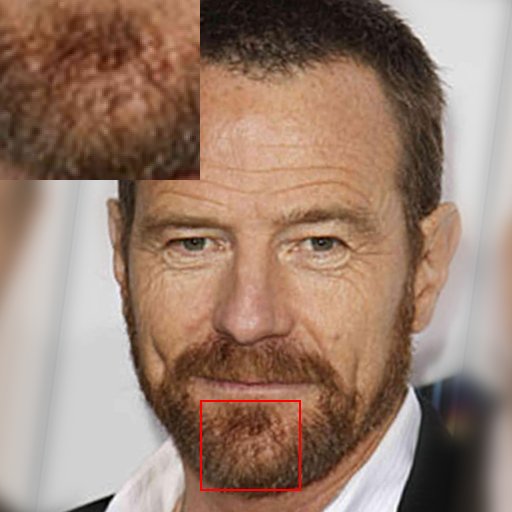} &
        \includegraphics[width=0.24\linewidth]{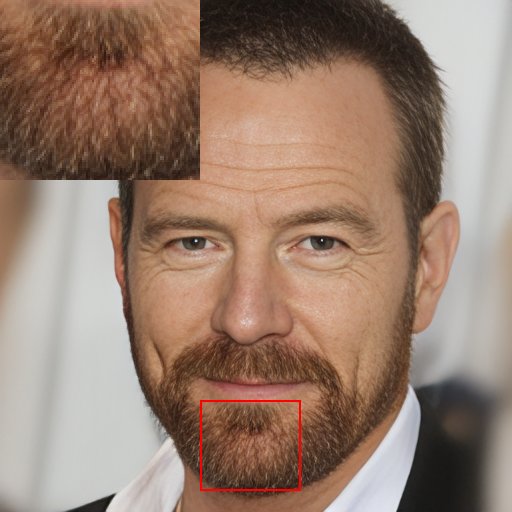} &
        \includegraphics[width=0.24\linewidth]{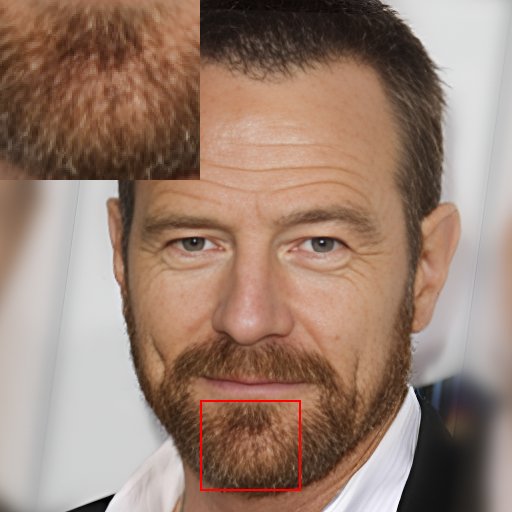} & 
        \includegraphics[width=0.24\linewidth]{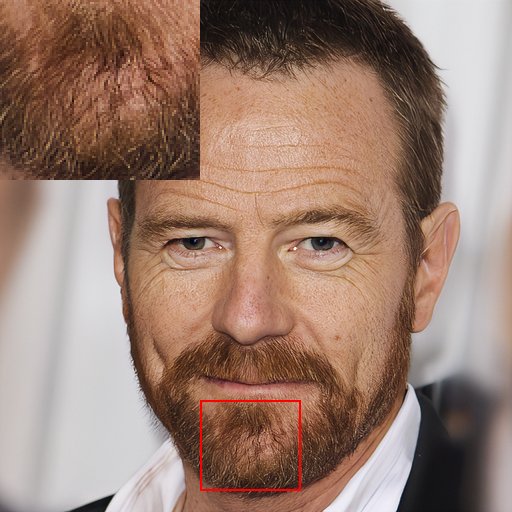}\\
        \includegraphics[width=0.24\linewidth]{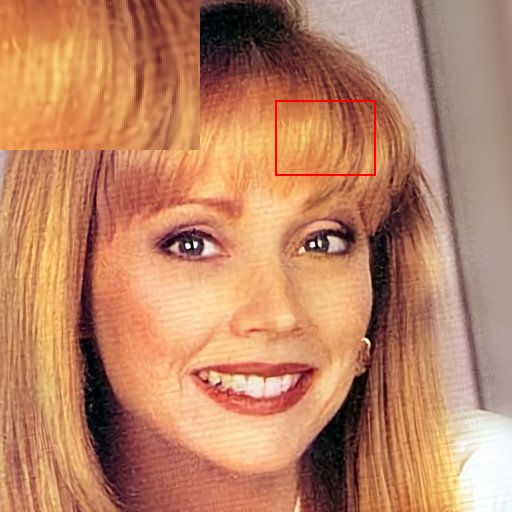} &
        \includegraphics[width=0.24\linewidth]{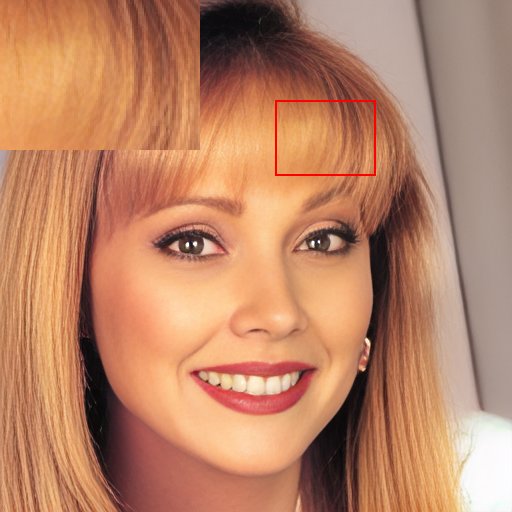} &
        \includegraphics[width=0.24\linewidth]{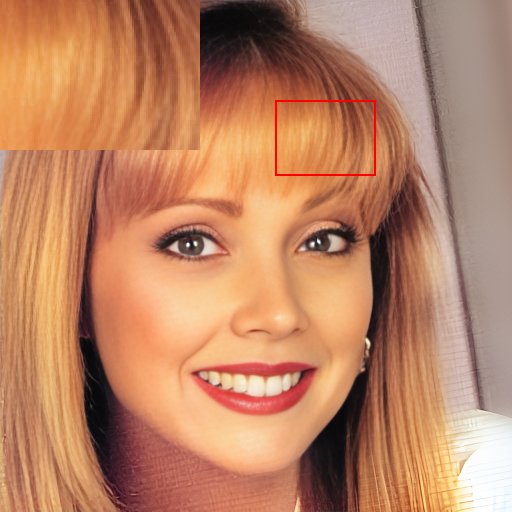} & 
        \includegraphics[width=0.24\linewidth]{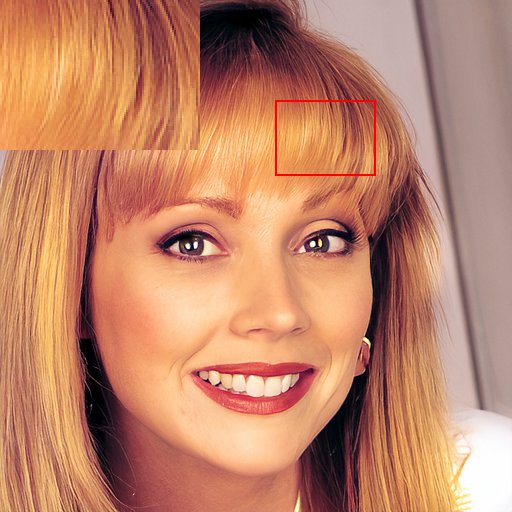}\\ 
        \bottomrule
    \end{tabular}
\endgroup
}
    \caption{Qualitative comparison on original CelebA-HQ test examples (without degradations). Baselines are biased by losing face details and generating hazy facial features. Our method authentically restores faces and preserves details much better regardless of the input quality. Zoom in to see richer details.}
    \label{fig:authentic_restore}
    \vspace{-9pt}
\end{figure*}

Face images convey critical information and are used in every corner of our daily lives, \eg, social networks, conference calls, and taking portrait, to name a few. Humans are very sensitive to delicate facial features. High-quality faces are the most preferred and typically are the key to the success of these applications. However, unknown and complicated image degradations are inevitable in real-world scenarios due to \eg camera defocus, encoding artifacts, motion blur, and noises. In recent years, rapid progress has been made in developing a \textit{blind face restoration} capability as a post-processing step to restore low-quality photos into high-quality ones faithfully.

Most existing popular works exploit pretrained facial prior networks~\cite{wang2021towards,zhou2022towards,chan2021glean,yang2021gan,glean_pami,chan2021glean} in assisting recovering photorealistic facial features even if the input is much deficient. Their success derived from the advanced power of generative models in encapsulating high quality priors, \eg, StyleGANs~\cite{goodfellow2014generative,karras2020analyzing} and VQVAE~\cite{van2017neural,razavi2019generating}. Other works use auxiliary information as guidance, such as face landmarks~\cite{chen2018fsrnet}, component features~\cite{li2020blind}, and high-quality exemplars~\cite{li2020enhanced,li2018learning}. Advanced architectures, including CNNs and Transformers~\cite{wang2022restoreformer,liang2021swinir} have been explored to alleviate face degradation issues. These works are based on building U-net models that restore low-quality faces into high-quality counterparts. The performance difference is often primarily attributed to the model design. 

Although these works have achieved notable progress by generating perceptually realistic faces and reporting good numeric results on the benchmarks like PSNR (Peak Signal-to-Noise Ratio) and FID (Fréchet Inception Distance)~\cite{heusel2017gans}, they still struggle over several important factors, e.g., fail to preserve delicate identity features, hallucinate uncanny artifacts especially in the presence of severe degradation, lose desired high-frequency details such as around facial hair. A quick test is to let the model perform restoration on slightly degraded faces that can make us distinguish delicate differences before and after restoration. As shown in Figure~\ref{fig:authentic_restore}, we evaluate two state-of-the-art works CodeFormer~\cite{zhou2022towards} and GFPGAN~\cite{wang2021towards} on four examples from CelebA-HQ data~\cite{karras2018progressive,liu2015faceattributes}. Both models either lose identity details, \eg, freckles and eyelashes or unexpectedly damage the original high-quality components like mustache and hair. The former issue sometimes raises ethical concerns, and the latter holds users from applying the restoration model broadly to fulfill their demands as the regression is unsatisfactory. We conjecture the aforementioned limitations not only from the design choices of their model, but also the quality of training data, as some ground-truth images are still noisy as seen in Figure~\ref{fig:authentic_restore}. In some sense, it contradicts the goal of recovering low-quality images into high-quality images if the ground-truth data is not clean, implying that training a restoration model is an ill-posed problem. Moreover, calculating pairwise metrics like PSNR between the predictions and the ground truth could also be problematic in evaluation. Intuitively, noisy training data can harm the restoration model's performance compared with clean data.

\vspace*{-0.1in}
\paragraph{Questions}
To address above challenges, there are two main questions to answer, which are also the primary factors that should be considered when building a machine learning system. 
$(\RN{1})$
\textbf{Model}: When processing above examples, why current model designs are unsatisfactory in balancing degradation removal and detail refinement?
$(\RN{2})$
\textbf{Data}: Can we automatically clean the data without expensive labor annotation and pave the path for authentic restoration and accurate evaluation?

\vspace*{-0.1in}
\paragraph{Solution}
To this end, we propose to use conditional denoising diffusion models (DDMs)~\cite{ho2020denoising} with well-designed iterative learning as a unified solution to both questions. Because of the inspiring and unprecedented results from image synthesis \eg, text-to-image synthesis~\cite{rombach2022high,saharia2022photorealistic} and image editing~\cite{ruiz2022dreambooth}, DDMs have attracted more and more attention from research community and shown great potentials to land in industry applications. There also have been works exploring DDMs for image restoration~\cite{kawar2022denoising,yin2022diffgar,saharia2022palette}, however, none of them yields a solution to address the above concerns. In this paper, the proposed iterative DDMs have the following properties:

\noindent $(\RN{1})$
\textit{Intrinsic iterative refinement}: DDMs compromise of iterative Markov chains in both forward diffusion and reverse denoising. The very dense U-Net architecture design and iterative refinement procedure make it a better choice to naturally preserve high-quality contents and remove degradation patterns.

\noindent  $(\RN{2})$
\textit{Extrinsic iterative enhancement}: As shown in Figure~\ref{fig:authentic_restore}, DDMs are found to balance degradation removal and detail preservation very well. Hence, the DDMs trained for restoration are ready for enhancing the training data which is then used for the next iteration of restoration training and more accurate evaluation.

\begin{figure}[t]
    \centering
    \begin{tabular}{cc}
    \toprule
    Source & Restored \\ \midrule
        \includegraphics[width=0.42\linewidth]{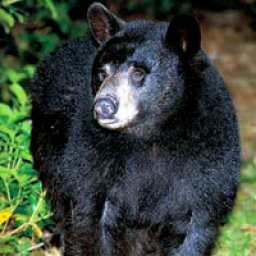} & \includegraphics[width=0.42\linewidth]{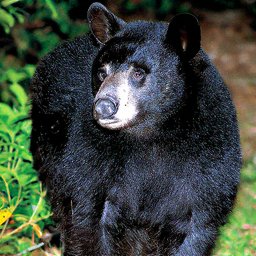} \\
        \bottomrule
    \end{tabular}
    \caption{Restoring original ImageNet $256 \times 256$ image with the proposed iterative DDMs. (Zoom in to see differences before and after.)}
    \label{fig:imagenet_restore}
    \vspace{-12pt}
\end{figure}

\vspace{2mm} \noindent \textbf{Beyond restoration}
Besides, we would rather think about using a model elsewhere than just for a specific goal, here, restoration. If above challenges are well resolved, we are able to get a new copy of higher-quality data in terms of resolution and cleanness. This consequently benefits generative models for training stabilization by reducing the number of outliers in terms of data quality and generating higher-fidelity samples. For example, as is known to all, the resolution of ImageNet can be as small as $75 \times 56$ and encoding compression is also noticeable in many images. Figure~\ref{fig:imagenet_restore} shows qualitative restored examples using a model trained on ImageNet restoration. Consistently, the sample quality has been increased a lot.  More inspiring results from generative models will be detailed in later sections.

\vspace{2mm} \noindent \textbf{Contribution}
The main contributions of this paper are as follows.
First, we point out the issues of existing face restoration systems and formulate the definition of authentic restoration that can quickly examine the effectiveness of any face restoration model.
Second, we build an iterative diffusion model (IDM) to tackle the raised concerns.
Finally, we demonstrate strong empirical results on two benchmarks, blind face restoration and image generation, showing that the proposed model consistently outperforms state-of-the-art methods and can be used to benefit the learning of generative models beyond restoration.

\section{Related Works}
\label{sec:related_works}

\paragraph{Image restoration}
Due to the presence of various image degradations, image restoration has attracted considerable attention from multiple aspects, \eg, deblurring~\cite{yasarla2020deblurring,shen2018deep,kupyn2019deblurgan}, denoising~\cite{zhang2018ffdnet,guo2019toward} and super-resolution~\cite{guo2017deep,wang2018esrgan,menon2020pulse,yang2020hifacegan,ma2020deep}. Among these topics, blind face restoration~\cite{li2018learning,li2020blind,wang2021towards,yang2021gan} (BFR) is of particular interest to us. Most approaches fall into one stage of learning where a U-Net model directly outputs the prediction given the input. In this category, utilizing the pretrained facial prior based on GANs and VAEs is the most commonly used strategy to handle a wide range of complicated degradations~\cite{wang2021towards,zhou2022towards,chan2021glean,yang2021gan}. This line of works assume the model can presumably produce realistic faces by projecting the low-quality faces into a compact low-dimension space of the pretrained generator. However, these models are suboptimal by hallucinating input objects and damaging high-quality details, as the best GAN inversion method cannot guarantee a nearly same reconstruction even for the high quality inputs. In contrast, our work is trained from scratch and can do an authentic restoration.

There are works exploring multi-stage learning by decomposing the task into several more tractable subtasks that refine the output in a coarse-to-fine manner and show improved performances~\cite{chen2021progressive,zamir2021multi,nah2017deep}. Depending on the model design, each subtask may output coarse feature or image predictions to the next stage. Our work is mainly related to this category of work. But different from them, diffusion models' learning and model design can be viewed as a whole without the need for delicate control over losses and architectures.

\vspace{1mm} \noindent \textbf{Denoising diffusion models}
Denoising diffusion models (DDMs) though they have been tried for image restoration~\cite{kawar2022denoising,yin2022diffgar,saharia2022palette}, none of them has shown the potential to surpass current state-of-the-art face restoration works. To our best knowledge, it is the first work that presents a thorough study of using DDMs for authentic face restoration.

\section{Methodology}
\label{sec:method}

\begin{figure*}[!ht]
    \centering
    \includegraphics[width=0.95\linewidth]{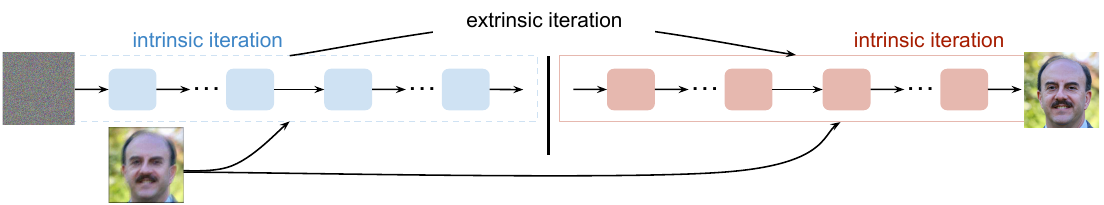}
    \caption{Illustration of intrinsic and extrinsic iterative learning. Both intrinsic iterations are conditioned on the synthetically degraded samples from original data. The extrinsic iteration consists of two intrinsic iterations of learning. The resulting model from the first intrinsic iteration is used to enhance the training data only which will then be used as the target data for the second intrinsic iteration.}
    \label{fig:model}
    \vspace{-9pt}
\end{figure*}

In this section, we start with the problem formulation of authentic face restoration. Next, we introduce the core iterative restoration method from two aspects, 
$(\RN{1})$ \textit{Intrinsic iterative learning} benefits authentic restoration via iterative refinement.
$(\RN{2})$ \textit{Extrinsic iterative learning} further improves restoration by automatically enhancing the training data.

\subsection{Problem Formulation}
Let $X$ denotes the high-quality image domain and $P_{X}$ implies the distribution of $X$. Assume that there exists a non-learnable one-to-many degradation function $d$ that maps $X$ to degraded low-quality image domain $X_d$: $X \rightarrow X_d$,
the goal of face restoration is to learn an inverse function $f_\theta$ parameterized by $\theta$: $X_d \rightarrow X$ that satisfies
\begin{equation}
    \min_{f_\theta} \mathcal{D}(P_{f_\theta(X_d)}||P_X) + \mathbb{E}_{x\sim X} \mathbb{E}_{x_d \sim f(x)} \kappa (f_\theta(x_d), x),
    \label{eq:face_restoration}
\end{equation}
where $\mathcal{D}$ is a distribution distance, \eg, KL (Kullback–Leibler) divergence (or maximum likelihood estimation)~\cite{kingma2013auto} or Jensen-Shannon divergence (or adversarial loss)~\cite{goodfellow2014generative}; $\kappa(\cdot)$ is an instance-wise distance between two images, \eg, perceptual loss~\cite{johnson2016perceptual} or $L_p$ distance ($p \sim \{1, 2\}$)~\cite{zhao2022rethinking}. All current face restoration works can be formulated by Eq.~\ref{eq:face_restoration}. For example, GFPGAN~\cite{wang2021towards} and CodeFormer~\cite{zhou2022towards} both keep the adversarial loss for high-fidelity restoration and mix different kinds of $\kappa(\cdot)$ for input's content preservation. In contrast, the proposed approach does not explicitly puts constraint to preserve content. Instead, the content preservation is achieved via injecting conditional signals inside the model plus iterative refinement. Thus, the second terms is omitted.

\vspace{-4mm}
\paragraph{Authentic restoration} In real applications, it is common that the degraded input $x_d \sim X_d$ can be either severely degraded, only has minor degradation, or even as clean as $x \sim X$. Since there is not a reliable metric to evaluate a single image's quality, \eg, sharpness and realisticness, an authentic restoration model is expected to treat $x$ and $x_d$ the same regardless the actual quality. Thus, we propose the following criterion that an authentic restoration model should obey:
\begin{equation}\label{eq:two_iter_authentic}
    f_\theta(x_d) = x, \quad f_\theta(x) = x
\end{equation}
which implies applying the model $f_\theta$ iteratively by starting from $x_d$ should converge to the same high-quality image $x$,
\begin{equation}\label{eq:authentic}
    x = f_\theta ... f_\theta(f_\theta(x_d))
\end{equation}
A mild relaxation is $f_\theta(f_\theta(x_d))$ is not worse than $x$ after two iterations. As is shown in Figure~\ref{fig:authentic_restore}, both baselines GFPGAN and CodeFormer fail the authentication test, \eg, hair detail is lost and freckles are removed. The results demonstrate that the baselines do not meet the criterion in Eq.~\ref{eq:two_iter_authentic}.

In this paper, by assuming the input is always of low quality, we can focus on two iterative restorations to deal with almost all real restoration scenarios. Next, we will describe how iterative diffusion models naturally couple with authentic restoration in Eq.~\ref{eq:authentic} with proper training.

\subsection{Intrinsic Iterative Learning}
Briefly, DDMs have a Markov chain structure residing in both the forward diffusion process and the reverse denoising chain. To recover the high-quality image, this structure naturally benefits authentic restoration via iterative refinement of the low-quality input, termed as \textit{intrinsic iterative learning}.
\vspace{-4mm}
\paragraph{Conditional DDMs}
DDMs are designed to learn a data distribution $q(x_0)$ by defining a chain of latent variable models $p_\theta(x_{0:T})=p(x_T)\: \Pi_{t=1}^T p_\theta(x_{t-1}|x_t)$,
\begin{equation}\label{eq:reverse_chain}
    x_T \rightarrow x_{T-1} \rightarrow ... \rightarrow x_1 \rightarrow x_0
\end{equation}
where each timestep's example $x_t$ has the same dimensionality $x_t \in \mathbb{R}^D$. Usually, the chain starts from a standard Gaussian distribution $x_T \sim \mathcal{N}(0, I^D)$ and only the final sample $x_0$ is stored. For restoration, we are more interested in the conditional DDMs that maps a low-quality source input $x_d$ to a high-quality target image $x$. The conditional DDMs iteratively edits the noisy intermediate image $x_t$ by learning a conditional distribution $p_\theta(x_{t-1} | x_t, x_d)$ such that $x_0 \sim p(x|x_d)$ where $x_0 \equiv x, x_d = d(x), \: x \sim P_X$.

\vspace{-4mm}
\paragraph{Learning} Let's first review the forward diffusion process that can be simplified into a specification of the true posterior distribution
\begin{equation}\label{eq:posterior}
    q(x_t|x_0) = \mathcal{N}(x_t|\sqrt{\gamma_t} x_0, (1 - \gamma_t) I)
\end{equation}
where $\gamma_t$ defines the noise schedule.

We thus learn the reverse chain with a ``denoiser" model $f_\theta$ which takes both source image $x_d$ and a intermediate noisy target image $x_t$ by comparing it $p_\theta(x_{t-1} | x_t, x_d)$ with the tractable posterior with $q(x_{t-1}|x_t, x_0)$. Consequently, following~\cite{saharia2022image}, we aim to optimize the following objective that estimates the noise $\epsilon$,
\begin{equation}\label{eq:loss}
    \mathcal{L} = \mathbb{E}_{(x, x_d)} \mathbb{E}_{\epsilon, \gamma} || f_\theta(x_d,  \hat{x}, \gamma)) - \epsilon)||_p^p, \: \epsilon \sim \mathcal{N}(0, 1).
\end{equation}
where $\hat{x} = \sqrt{\gamma} x_0 + \sqrt{(1 - \gamma)}\epsilon$ as in Eq.~\ref{eq:posterior} and usually we can set $p \in \{1, 2\}$. One can also directly predict the output of $f_\theta$ to $x_0$, a.k.a, regression, which makes \eqref{eq:loss} becomes:
\begin{equation}\label{eq:loss_new}
    \mathcal{L} = \mathbb{E}_{(x, x_d)} \mathbb{E}_{\gamma} || f_\theta(x_d,  \hat{x}, \gamma)) - x_0)||_p^p.
\end{equation}
We find this formulation more efficient in both training and inference since the sequence of network approximates $x_0$ at each time step starting from different amount of noises. This makes our work differ from previous conditional image-to-image synthesis works SR3~\cite{saharia2022image} and  Palette~\cite{saharia2022palette}.

\begin{figure}[t]
    \centering
    \includegraphics[width=0.95\linewidth]{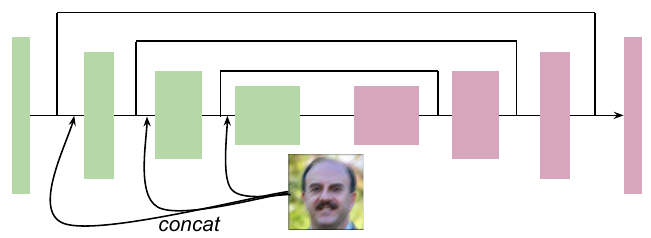}
    \caption{Dense U-Net with conditional signals.}
    \label{fig:unet}
    \vspace{-12pt}
\end{figure}

\paragraph{Iterative restoration}
As for the reverse diffusion process during the inference stage, we follow the Langevin dynamics~\cite{saharia2022image},
\begin{equation}\label{eq:langevin}
    x_{t-1} = \frac{1}{\sqrt{\alpha_t}} \left(x_t - \frac{1-\alpha_t}{\sqrt{1-\gamma_t}}f_\theta(x_t, x_d, \gamma_t) \right) + \sqrt{1 - \alpha_t} \epsilon_t
\end{equation}
where $\alpha_t$ is hyper-parameter related to $\gamma_t=\Pi_{i=1}^t \alpha_i$~\cite{saharia2022image}.
To help understand why DDMs can provide authentic restoration, we justify from two aspects,
\begin{itemize}[leftmargin=*,label=$\bullet$,topsep=2pt]
    \setlength{\itemsep}{2pt}
    \setlength{\parskip}{2pt}
    \item \textbf{Iterative refinement}: the above process iteratively refines the input towards the target high-quality image. Different time steps are guided to learn restoration with an annealing noise schedule. That is, it coincides with our authentic restoration motivation in Eq.~\ref{eq:authentic}. Theoretically, as $\alpha_t$ is annealed to 1, it makes the sufficiently long denoising chain always converge to the same data point, that is,
    \begin{equation}
        x_t \approx x_{t-1}
    \end{equation}
    \item \textbf{Dense architecture}: in convention as shown in Figure~\ref{fig:unet}, the design of the predictor $f_\theta$ is a very dense structure with hierarchical links between encoder and decoder. Conditional signal can also be strong to preserve the high-quality contents without any auxiliary losses. That is, $x_t$ will gradually learn to absorb the clean content from $x_d$ to approximate a clean version $x_0$ of $x_d$ as in Eq.\eqref{eq:loss_new}.
\end{itemize}

\subsection{Extrinsic Iterative Learning}
Most works choose to use 70K FFHQ (Flickr-Faces-HQ) data~\cite{karras2020analyzing} as training data. The dataset is currently the largest public high-quality high-resolution face data and is assumed to be clean and sharp. It is collected from internet and filtered by both automatic annotation tools and human workers. However, revisiting the examples in Figure~\ref{fig:authentic_restore} reveals that the training data is not always high-quality. JPEG degradation, blur and Gaussian noise are found to account for around 15\% when we examine the first 2000 images. The quality of dataset has proven to be a key factor in learning a restoration model and we will also provide an ablation later. To avoid being trapped in such a dilemma, we herein propose a simple yet effective solution  \textit{extrinsic iterative learning}. Thanks to the capability of authentic restoration of DDMs, the solution can automatically restore the training data without 
damaging the data, especially for those whose quality are already high and facial details are ineligible.

\paragraph{Iterative restoration}
As shown in Figure~\ref{fig:authentic_restore}, DDMs are proven to satisfy  Eq.~\ref{eq:two_iter_authentic}.  After learning the restoration model $f_\theta$, we can apply it to all the training data to produce a new high-quality image domain $P_{X^*}$,
\begin{equation}\label{eq:extrinsic_restore}
    x^* = f_\theta(x), \quad x \sim P_X.
\end{equation}
We term the method as extrinsic iterative learning because Eq.\ref{eq:extrinsic_restore} produces a higher quality data for another iteration of restoration model training, which is different from the internal chained process in DDMs.

Note that we still draw low-quality image samples $x_d$ with mapping: $X \rightarrow X_d$ and the target distribution becomes $P_{X^*}$ instead of $P_X$. Therefore we learn a conditional distribution $p_{\theta^*}(x_{t-1}|x_t, x_d)$ such that $x_0 \sim p(x^*|x_d)$ where $x_0 \equiv x^*, \: x_d = d(x), \: x \sim P_X$.

\paragraph{Coupling two DDMs}
Note that the only difference of learning $f_\theta \:\text{and}\: f_{\theta^*}$ is the target distribution $P_X$ and $P_{X^*}$. With proper tuning, the pool of $P_X$ can be gradually replaced and filled by new data from $P_{X^*}$. Given an ideal ``stop sign" for learning $f_\theta$, we actually can unify the two DDMs $f_\theta \:\text{and}\: f_{\theta^*}$ into only one learning process. In this paper, we do not adopt this option as we find simply training two separate models works well and saves parameter tuning cost. Nevertheless, we still can view the decoupled DDMs as a single model experimentally since $f_{\theta^*}$ is initialized from $f_\theta$. Through intrinsic and extrinsic iterative learning, we only keep the latest model $f_{\theta^*}$ for inference. 

\section{Experiments}
\label{sec:experiments}
\begin{figure*}[!ht]
    \centering
    \resizebox{0.95\linewidth}{!}{
    \setlength{\tabcolsep}{1pt}
    \renewcommand{\arraystretch}{0.8}
    \begin{tabular}{c|ccccc}
    \toprule
        Source & GFPGAN~\cite{wang2021towards} & CodeFormer~\cite{zhou2022towards} & VQFR~\cite{gu2022vqfr} & Ours & GT \\ \midrule
        \includegraphics[width=0.18\linewidth]{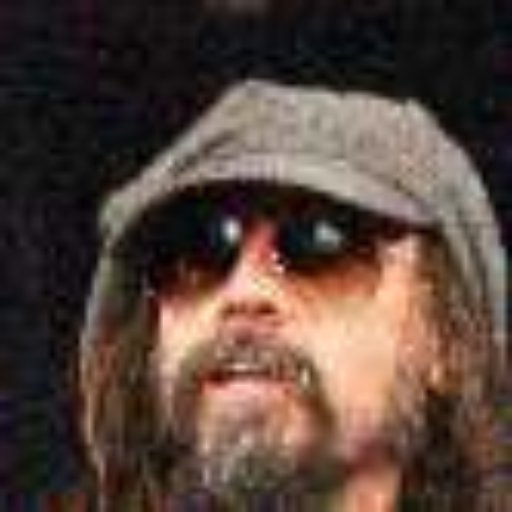} &
        \includegraphics[width=0.18\linewidth]{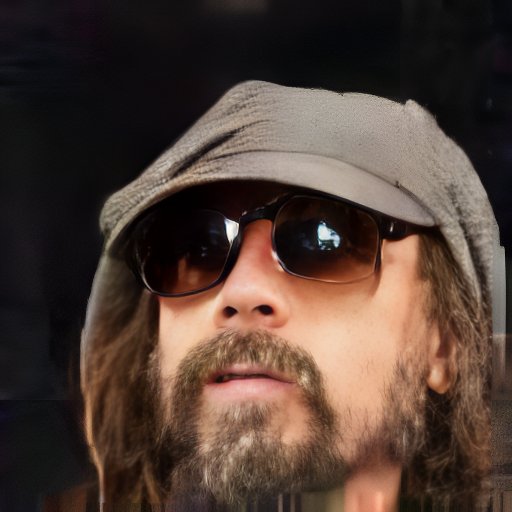} &
        \includegraphics[width=0.18\linewidth]{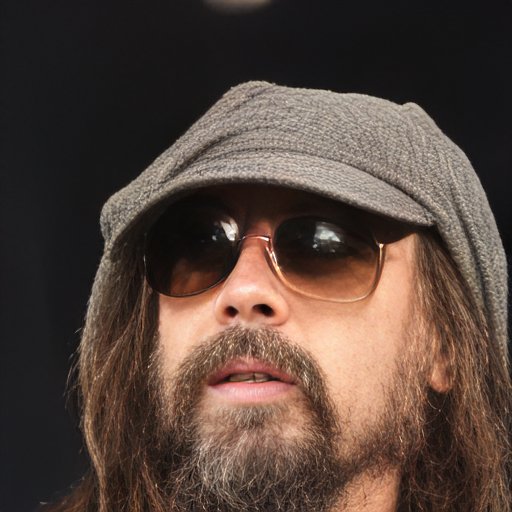}&
        \includegraphics[width=0.18\linewidth]{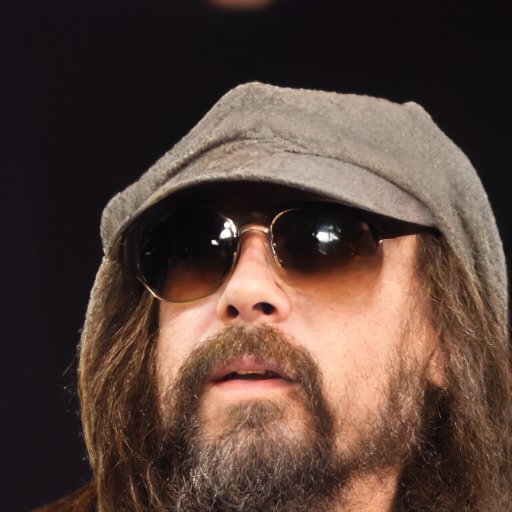}&
        \includegraphics[width=0.18\linewidth]{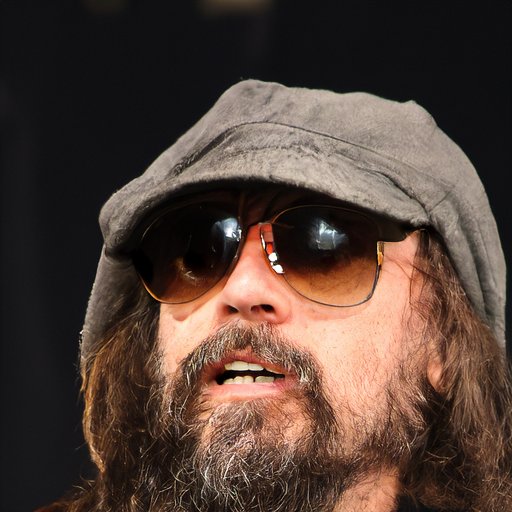}  & \includegraphics[width=0.18\linewidth]{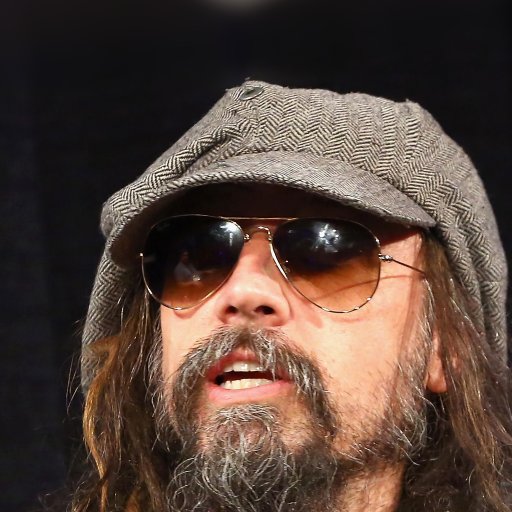} \\
        \includegraphics[width=0.18\linewidth]{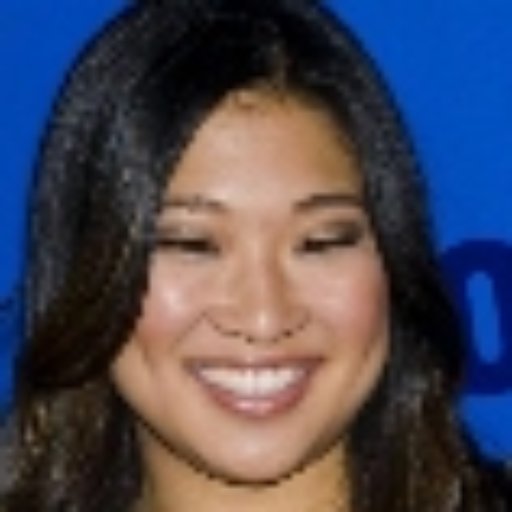} &
        \includegraphics[width=0.18\linewidth]{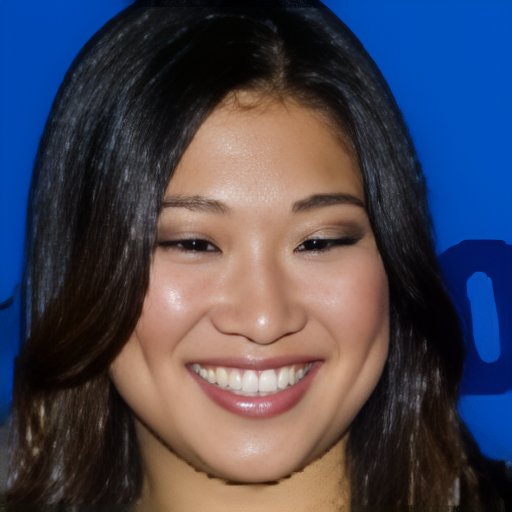}& 
        \includegraphics[width=0.18\linewidth]{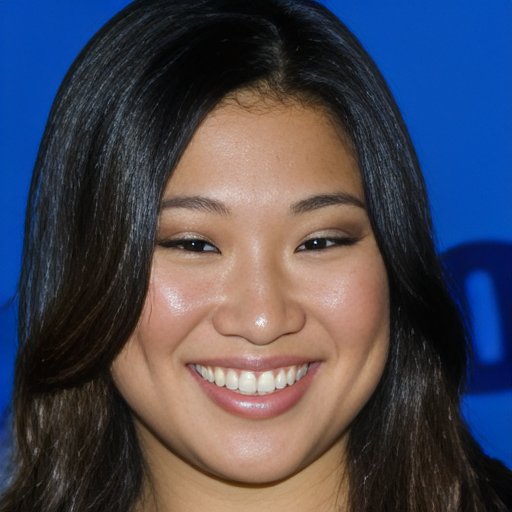}&
        \includegraphics[width=0.18\linewidth]{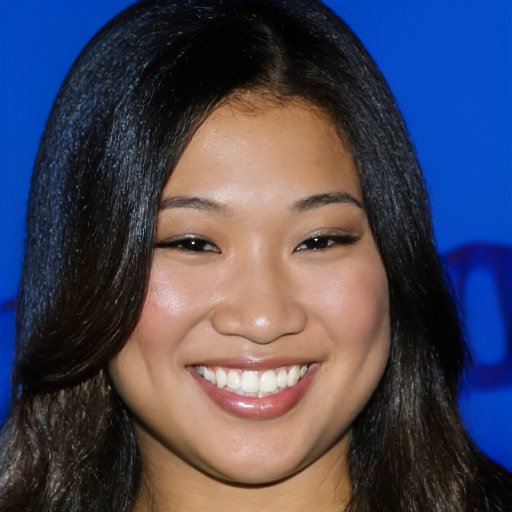}&
        \includegraphics[width=0.18\linewidth]{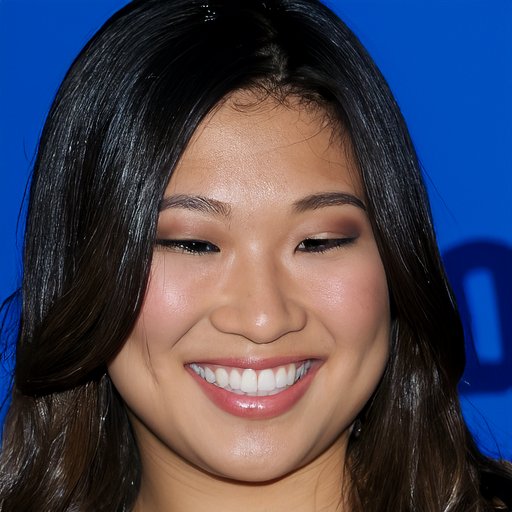} &  \includegraphics[width=0.18\linewidth]{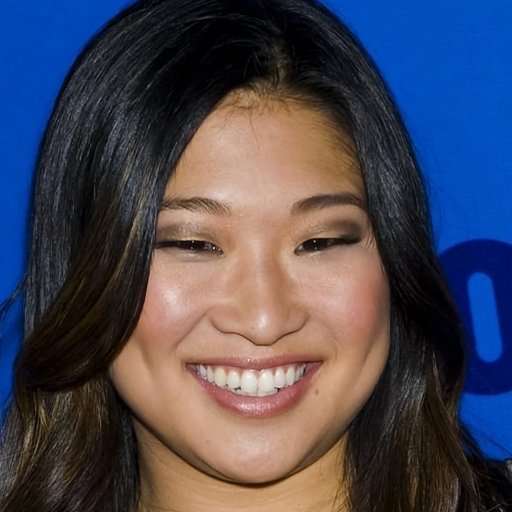}  \\ 
        \includegraphics[width=0.18\linewidth]{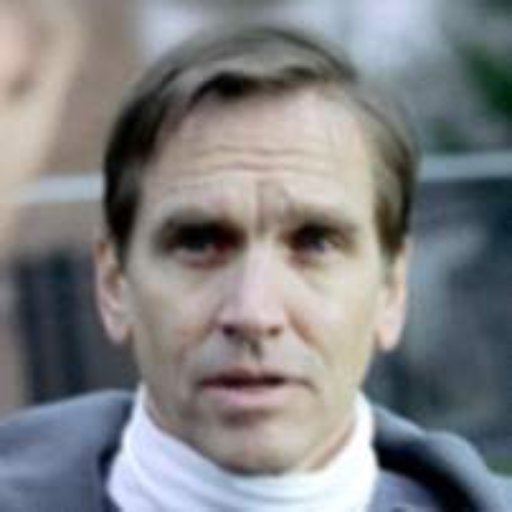} &
        \includegraphics[width=0.18\linewidth]{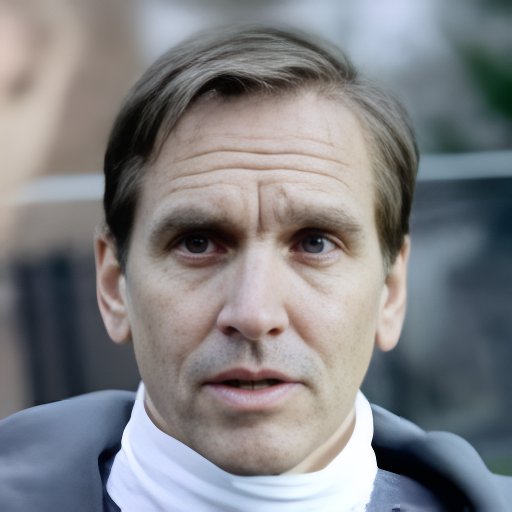}&
        \includegraphics[width=0.18\linewidth]{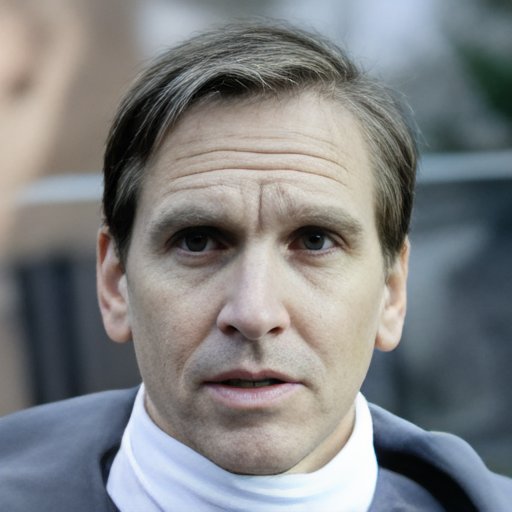} &
        \includegraphics[width=0.18\linewidth]{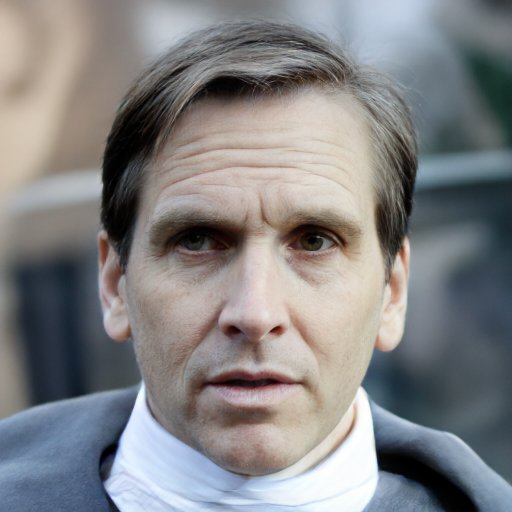}&
        \includegraphics[width=0.18\linewidth]{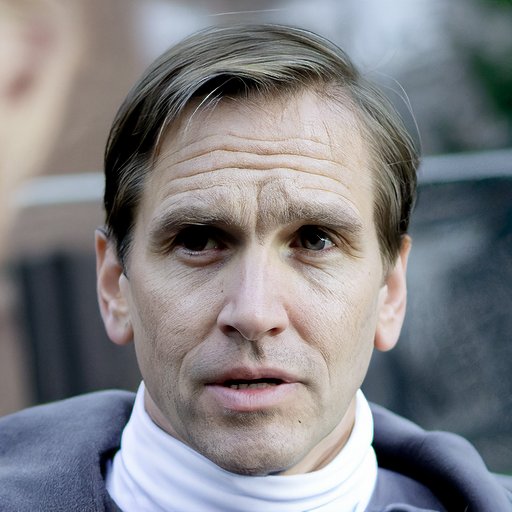}& \includegraphics[width=0.18\linewidth]{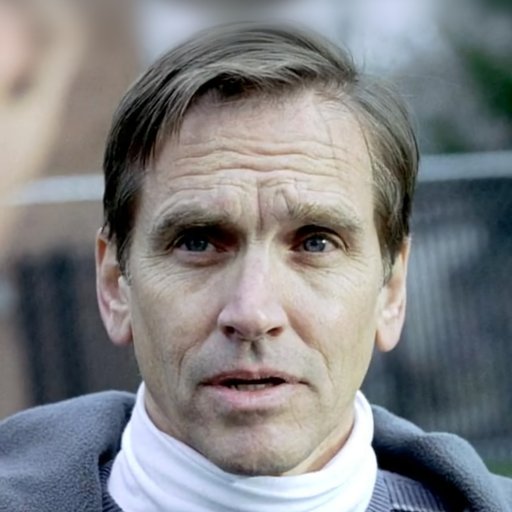} \\
    \bottomrule
    \end{tabular}
    }
    \caption{Qualitative comparison with state-of-the-art restoration models on CelebA-HQ Test. }
    \label{fig:celebehq_restore}
\end{figure*}

\begin{figure*}[!ht]
    \centering
    \resizebox{0.95\linewidth}{!}{
    \setlength{\tabcolsep}{1pt}
    \renewcommand{\arraystretch}{0.8}
    \begin{tabular}{c|cccc}
    \toprule
        Source & GFPGAN~\cite{wang2021towards} & CodeFormer~\cite{zhou2022towards} & VQFR~\cite{gu2022vqfr} & Ours\\ \midrule
        \includegraphics[width=0.18\linewidth]{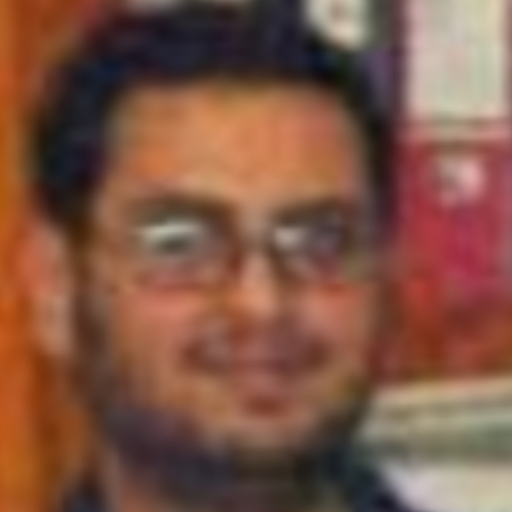} &
        \includegraphics[width=0.18\linewidth]{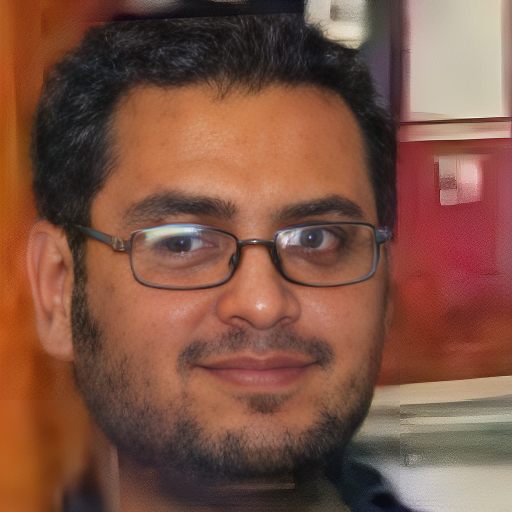} &
        \includegraphics[width=0.18\linewidth]{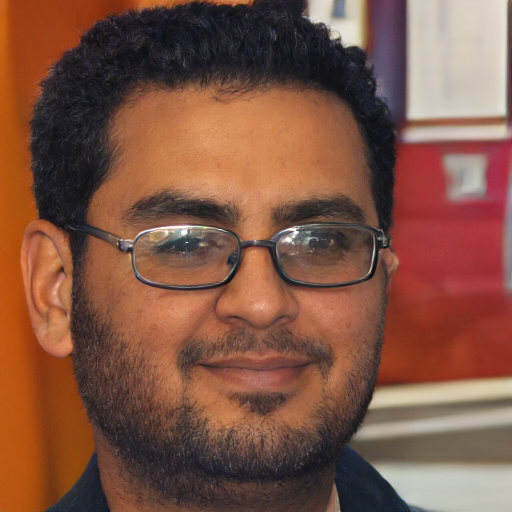}&
        \includegraphics[width=0.18\linewidth]{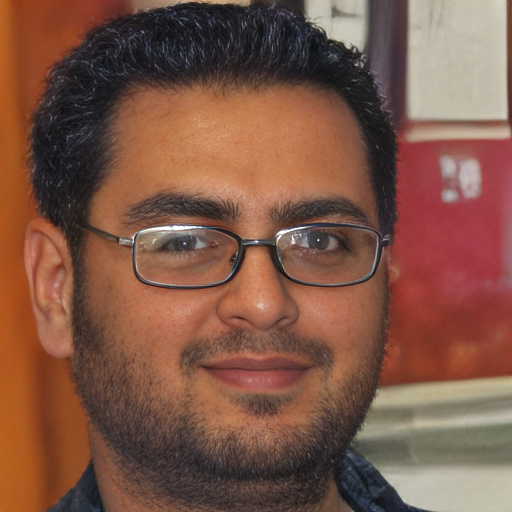}&
        \includegraphics[width=0.18\linewidth]{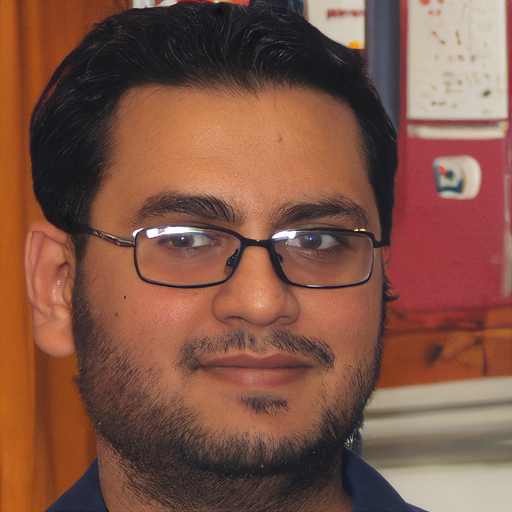} \\
        \includegraphics[width=0.18\linewidth]{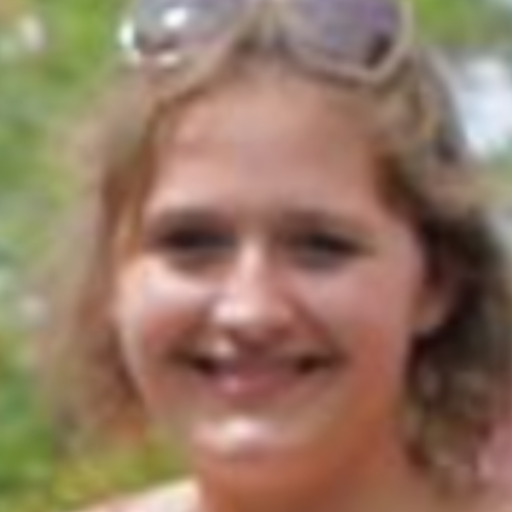} &
        \includegraphics[width=0.18\linewidth]{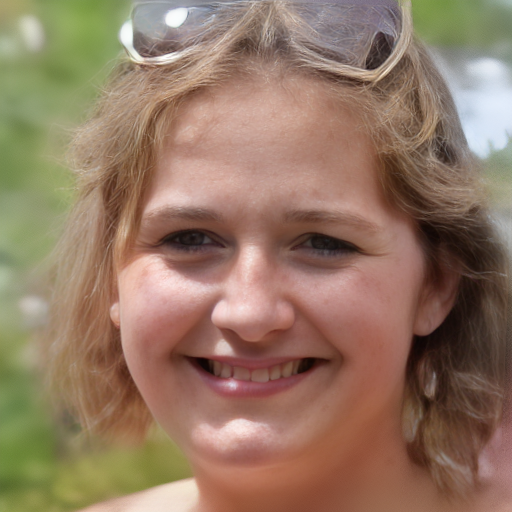} &
        \includegraphics[width=0.18\linewidth]{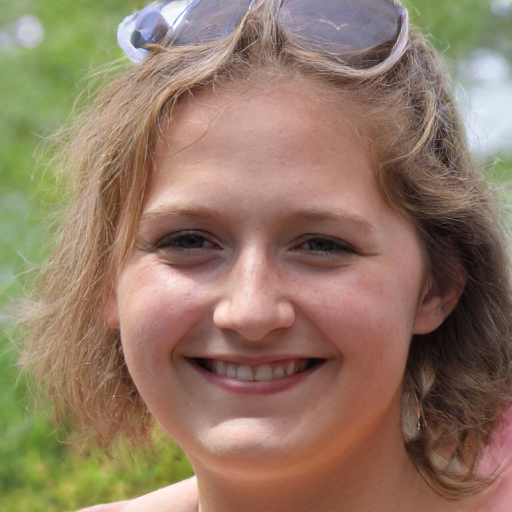}&
        \includegraphics[width=0.18\linewidth]{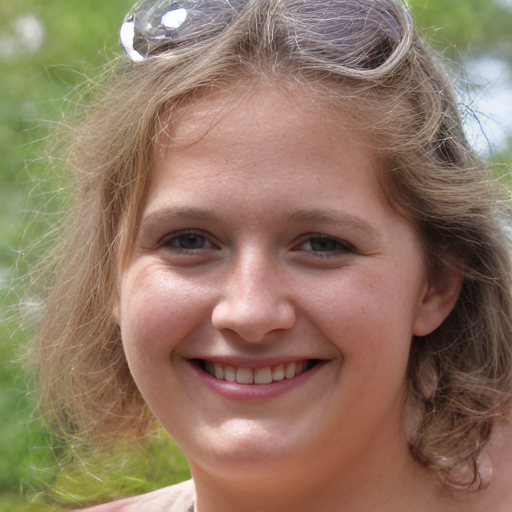}&
        \includegraphics[width=0.18\linewidth]{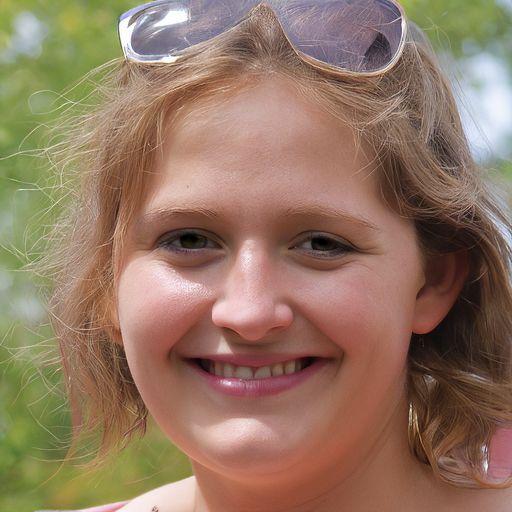} \\        
    \bottomrule
    \end{tabular}
    }
    \caption{Qualitative comparison with state-of-the-art restoration models on real-world Wider-Test datasets. }
    \label{fig:real_data_restore}
\end{figure*}

We first evaluate our method on blind face restoration. Next, we demonstrate that the proposed method can be used to improve other generation tasks. Finally, we conduct ablation studies on our model.

\subsection{Face restoration}

\paragraph{Dataset} 
We train our model on the 70K FFHQ dataset~\cite{karras2019style} and evaluate on the test split of the CelebA-HQ dataset~\cite{karras2018progressive}, which contains 30K images following GFPGAN\footnote{https://xinntao.github.io/projects/gfpgan}.
All images are resized to $512{\times}512$.
Following the standard practice in face restoration~\cite{li2018learning,wang2021towards,li2020blind,yang2021gan}, we synthesize degraded low-quality faces $x_d$ from real high-quality faces $x$ using the following degradation function:
\begin{equation} \label{eq:deg}
    x_d = \left[(x \otimes \mathbf{k}_\sigma)_{\downarrow_r} + \mathbf{n}_\delta \right]_{JPEG_q},
\end{equation}
\ie~the high-quality image $x$ is first convolved with a Gaussian blur kernel $\mathbf{k}_\sigma$ with kernel size $\sigma$ and downsampled by a factor $r$. Gaussian noise $\mathbf{n}_\delta$ with standard deviation $\delta$ is then added before applying JPEG compression with quality factor $q$ to obtain the final low-quality image $x_d$. Although there are possible other types of degradations, we adopt the same degradation model used in prior works~\cite{wang2021towards,li2020blind,yang2021gan} for a fair comparison. We randomly sample $\sigma$, $r$, $\delta$ and $q$ from [0.2, 10], [1, 8], [0, 15] and [50, 100] for the degradation function. Degradation pipeline follows GFPGAN~\cite{wang2021towards} implementation\footnote{https://github.com/TencentARC/GFPGAN}.
The restoration model is trained with image pairs $(x, x_d)$ following Section~\ref{sec:method}.

Additionally, as seen in Figure~\ref{fig:authentic_restore}, the proposed method has demonstrated the ability to authentically repair the data and CelebA-HQ is known to be pretty noisy, we also prepare a set of clear CelebA-HQ test set as new evaluation set. We will release ground-truth, degraded data and restored results from different models to reproduce. We also evaluate the proposed method on several real-world datasets, LFW, WebPhotos~\cite{wang2021towards} and Wider-Test~\cite{zhou2022towards}.

\paragraph{Baselines}
Three very recent models, GFPGAN~\cite{wang2021towards}, RestoreFormer~\cite{wang2022restoreformer} and CodeFormer~\cite{zhou2022towards}, are used as primary baselines. GFPGAN is the first successful approach that applies pretrained StyleGAN face prior for improved restoration. RestoreFormer explores transformer architecture as model base.  To approximate the high-quality face correspondence, CodeFormer predicts the latent codes with a transformer.

\paragraph{Metrics}
We mainly apply PSNR, SSIM, LPIPS~\cite{zhang2018unreasonable} and Arcface identity score (ID)~\cite{deng2019arcface} as quantitative evaluation metrics. In particular, we calculate LPIPS with VGG following official implementation\footnote{https://github.com/richzhang/PerceptualSimilarity}. 

\paragraph{Implementation details}
The DDM architecture follows that of~\cite{ho2020denoising}, except that we add attention layers in lower resolution ($32{\times}32$ to $8{\times}8$) blocks. Also, we directly train the model on $512{\times}512$ resolution instead of using the common cascade approach~\cite{ho2022cascaded,saharia2022image}. The model compromises of 120M parameters and takes about 2 days (100K iterations) to converge on eight V100 GPUs. We choose $T=1000$ for training and 10 diffusion steps for efficient inference.

\begin{figure}[!ht]
    \centering
    \begin{tabular}{cc}
        \includegraphics[width=0.45\linewidth]{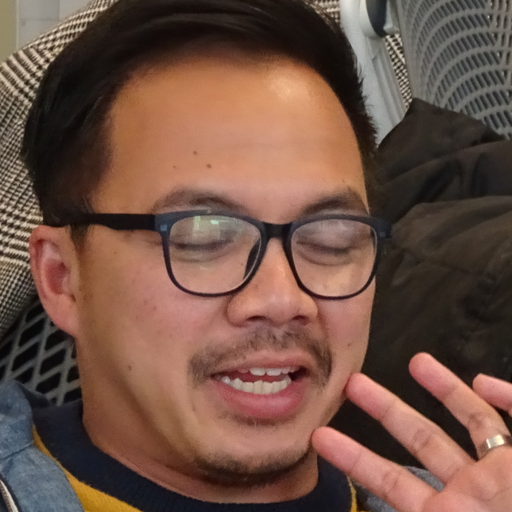} &
        \includegraphics[width=0.45\linewidth]{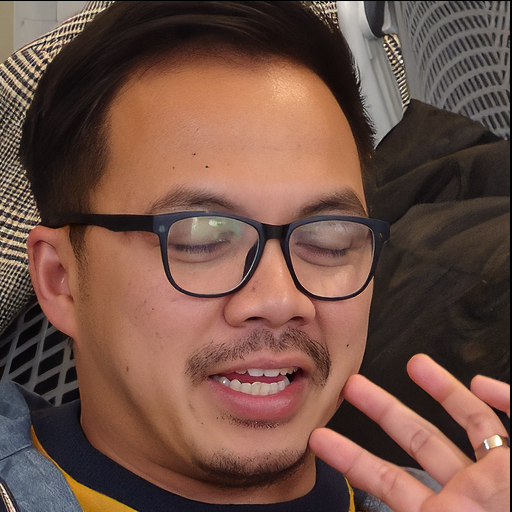} \\
    \end{tabular}
    \caption{An example from FFHQ. Before and after applying the first iteration restoration model.}
    \label{fig:ffhq_train_cmp}
\end{figure}

\paragraph{Results}
We first conduct an analysis on how significant extrinsic iterative learning can improve training dataset quality. As is shown in Figure~\ref{fig:ffhq_train_cmp}, the trained restoration model can make the data more clear while preserving all contents. We also compute the NIMA score on FFHQ~\cite{talebi2018nima}, commonly used to evaluate image quality without using reference. After restoration, the image quality of FFHQ dataset improves from 4.51 to 4.85.

\begin{table}[!ht]
    \centering
    \resizebox{0.9\linewidth}{!}{
    \centering
    \begin{tabular}{l|cccc}
    \toprule
        \textbf{Models}
        & PSNR$\uparrow$ & SSIM$\uparrow$ & ID$\uparrow$ & LPIPS$_\text{VGG}$$\downarrow$ \\ \midrule
        DFDNet~\cite{li2020blind} & 25.81 & 0.659 & 0.40 & 0.418\\
        GFPGAN~\cite{wang2021towards} & 26.95 & 0.715 & 0.59 & 0.359 \\
        ReFormer~\cite{wang2022restoreformer} & 25.83 & 0.671 & 0.68  & 0.360  \\ 
        VQFR~\cite{gu2022vqfr} & 24.60 & 0.631 & 0.70 & 0.341 \\
        CodeFormer~\cite{zhou2022towards} &  25.70 &  0.664 & 0.75  & 0.337  \\ \midrule
        IDM (Ours) & \textbf{27.42} & \textbf{0.742} & \textbf{0.87} & \textbf{0.307} \\
     \bottomrule
    \end{tabular}
    }
    \caption{Quantitative comparison on blind face restoration (BFR) using \textit{original} CelebA-HQ.
    }
    \label{tab:bfr-result}
\end{table}

\begin{table}[!ht]
    \centering
    \resizebox{0.9\linewidth}{!}{
    \centering
    \begin{tabular}{l|cccc}
    \toprule
        \textbf{Models}
        & PSNR$\uparrow$ & SSIM$\uparrow$ & ID$\uparrow$ & LPIPS$_\text{VGG}$$\downarrow$ \\ \midrule 
        DFDNet~\cite{li2020blind} & 24.10 & 0.621 & 0.39 & 0.404\\
        GFPGAN~\cite{wang2021towards} & 24.65 & 0.646 & 0.55 & 0.351 \\
        ReFormer~\cite{wang2022restoreformer} & 24.06 & 0.618 & 0.60  & 0.359  \\ 
        VQFR~\cite{gu2022vqfr} & 23.31 & 0.598 & 0.64 & 0.336 \\
        CodeFormer~\cite{zhou2022towards} &  23.89 &  0.613 & 0.79 & 0.314 \\ \midrule
        IDM (Ours) & \textbf{28.25} & \textbf{0.803} & \textbf{0.88} & \textbf{0.291} \\
     \bottomrule
    \end{tabular}
    }
    \caption{Quantitative comparison on blind face restoration (BFR) using \textit{clean} CelebA-HQ.
    }
    \label{tab:clean-bfr-result}
\end{table}

Table~\ref{tab:bfr-result} and Table~\ref{tab:clean-bfr-result} respectively show the results on the original and clean CelebA-HQ test set with randomly synthetic degradation. Our method consistently outperforms baselines across different metrics. This verifies the effectiveness of the proposed method on face restoration. In Figure~\ref{fig:celebehq_restore} and Figure~\ref{fig:real_data_restore}, comparing our method and baselines on both synthetic data and real-world data, it is observed that both baselines tend to generate hazy and unnatural faces. In contrast, our approach can give much more natural and realistic results with rich details. 

We also conduct an user study that collects users' opinions. We ask 10 raters to do a blind model evaluation on restoration results in clear CelebA-HQ test with synthetic degradations. Each rater is asked to select which model shows the best perceptual quality while preserving the facial details given the ground-truth images. As in Table~\ref{tab:bfr-user-result}, our approach outperforms baselines by a significant margin. 

\begin{table}[!ht]
    \centering
    \resizebox{1.0\linewidth}{!}{
    \centering
    \begin{tabular}{l|cccc|c}
    \toprule
        \textbf{Models} &  GFPGAN~\cite{wang2021towards}  & 
        ReFormer~\cite{wang2022restoreformer} &
        VQFR~\cite{gu2022vqfr} &
        CodeFormer~\cite{zhou2022towards} & 
        IDM (Ours)   \\ \midrule
        Preferences$\uparrow$ & 4.1$\%$ & 4.7$\%$ & 8.4$\%$ & 11.2$\%$ & 71.6$\%$ \\ 
     \bottomrule
    \end{tabular}
    }
    \caption{Subjective study on blind face restoration (BFR).
    }
    \label{tab:bfr-user-result}
\end{table}

\begin{figure*}[!ht]
    \centering
    \scalebox{0.95}{
    \begin{tabular}{cc|c}
    \toprule
    \multicolumn{2}{c|}{Uncurated FFHQ $512\times512$} & ImageNet $128\times128$\\ \midrule
    Baseline (FID=3.08) & Ours (FID=2.78) & \\
    \includegraphics[width=0.325\linewidth]{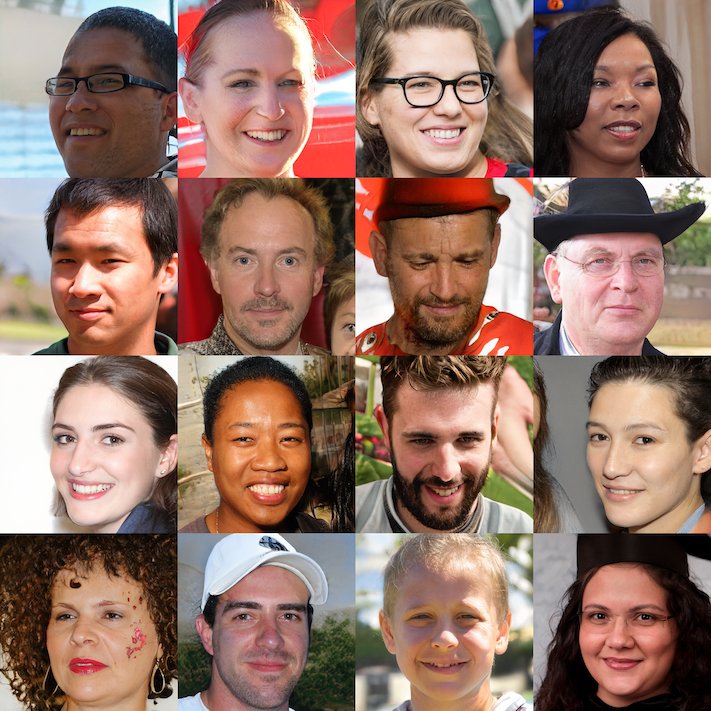} & \includegraphics[width=0.325\linewidth]{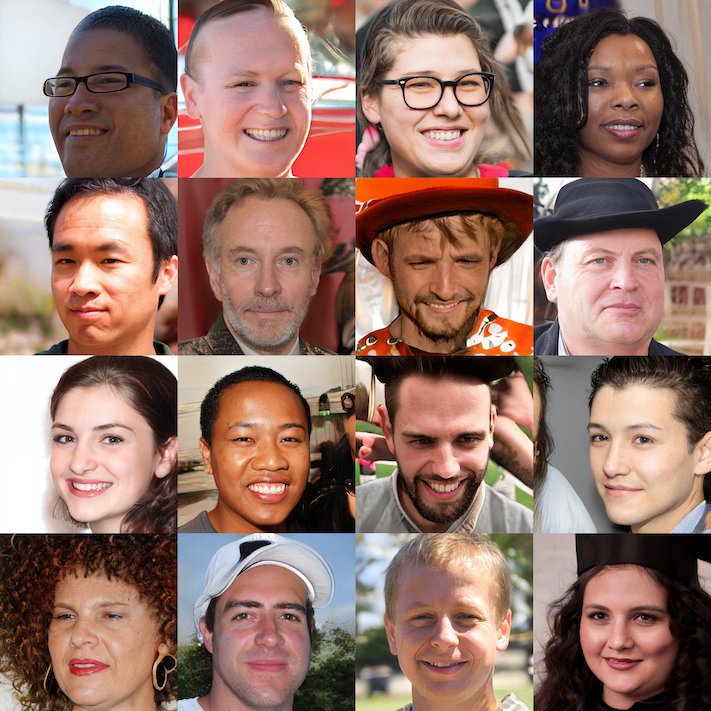} &
    \includegraphics[width=0.25\linewidth]{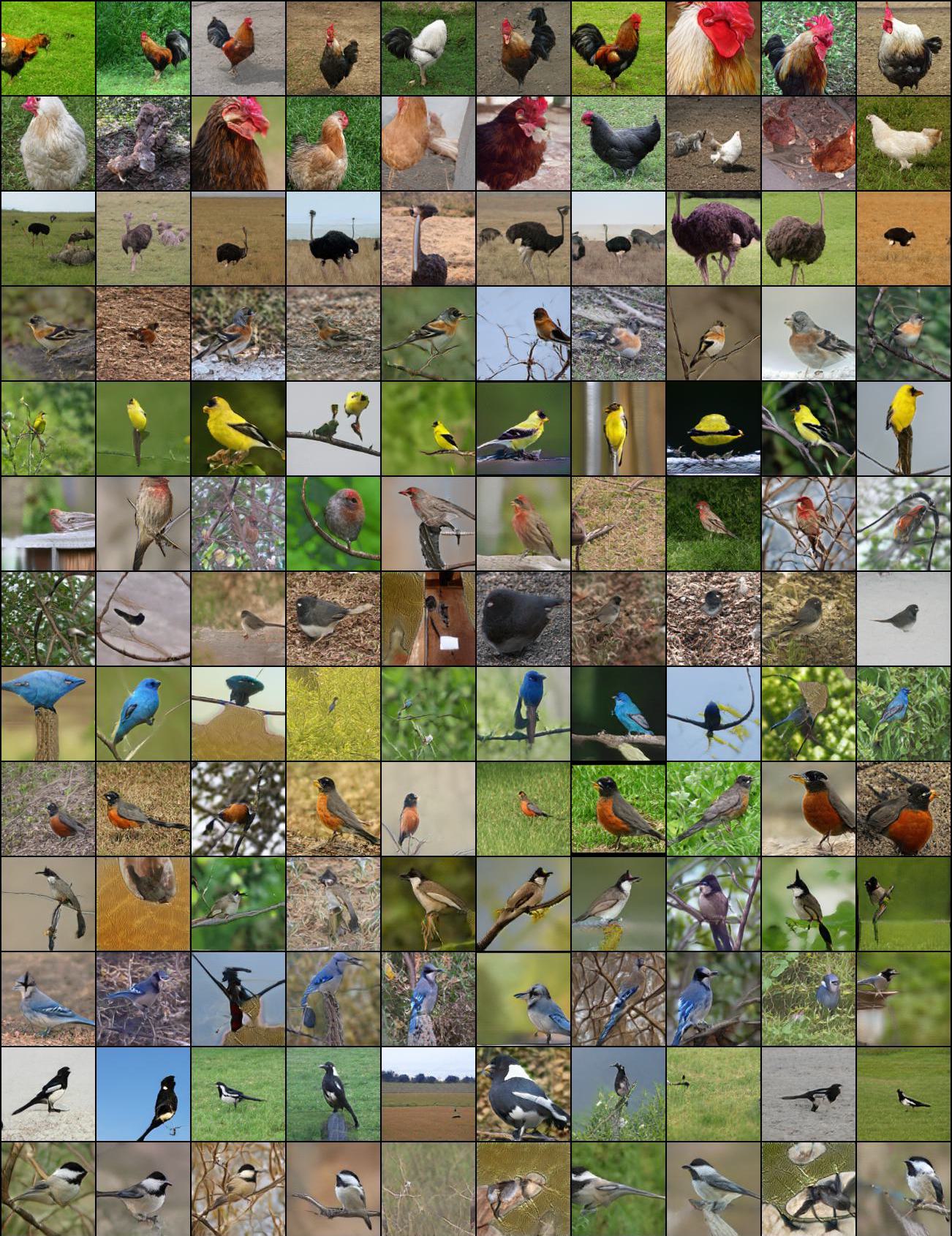}\\ \bottomrule
    \end{tabular}
    }
    \vspace{-3pt}
    \caption{Qualitative generated samples of FFHQ $512\times512$ from StyleGAN2 and ImageNet $128\times128$ from BigGAN.}
    \label{fig:samples_generative}
    \vspace{-6pt}
\end{figure*}

\subsection{Improved generation}
\begin{figure}[t]
    \centering
    \vspace{-12pt}
   \includegraphics[width=0.65\linewidth]{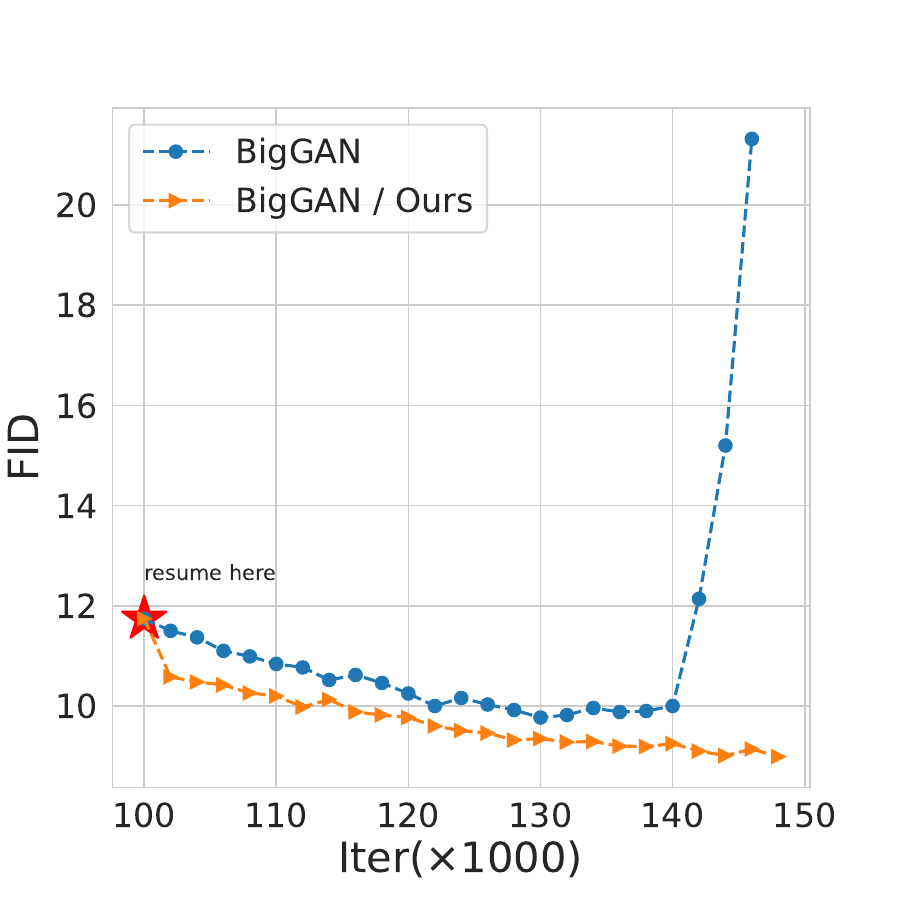}
    \caption{BigGAN training is stabilized on ImageNet $128\times128$.}
    \label{fig:biggan_imagenet}
    \vspace{-9pt}
\end{figure}

We also analyze the impact of restored dataset on generative models' performance. We adopt two most popular image generation benchmarks, FFHQ and ImageNet, and evaluate the method on two different kinds of models, GANs and DDMs. For FFHQ generation, we use official StyleGAN2-PyTorch implementation and resume training from the released checkpoint. For ImageNet with BigGAN, we use the official PyTorch BigGAN implementation\footnote{https://github.com/ajbrock/BigGAN-PyTorch} and resume the training from the suggested checkpoint for finetuning.  Numbers from BigGAN are reported as it is without truncation. For DDMs, we train the conditional DDMs on resolution $384\times384$ by conditioning on ImageNet-1k class labels. Architecture resembles the one being used in restoration model training. Quantitative results of FFHQ and ImageNet are respectively shown in Table~\ref{tab:ffhq_gen} and Table~\ref{tab:imagenet_gen}. The improved FID, precision and recall consistently suggest that refined data by the restoration model can help generative models to learn the data better and obtain higher-quality samples. Besides, in Figure~\ref{fig:biggan_imagenet}, it is shown that the training stability of GANs is also relieved a bit whereas the baseline BigGAN diverges after 140K iterations. We present qualitative generated samples in Figure~\ref{fig:samples_generative}. When comparing uncurated FFHQ samples for StyleGAN, we can see artifacts have been reduced.

\begin{table}[!ht]
    \centering
    \resizebox{0.9\linewidth}{!}{
    \begin{tabular}{c|l|ccc}
    \toprule
       Resolution & Methods & FID$\downarrow$ & Precision$\uparrow$ & Recall$\uparrow$ \\ \midrule
        \multirow{2}{*}{$\times 512$} & StyleGAN2~\cite{karras2020analyzing} &3.08 & 0.676 & 0.478 \\
        & IDM (Ours) & \textbf{2.78} & \textbf{0.683} & \textbf{0.479} \\ \midrule
        \multirow{6}{*}{$\times 256$} & U-Net GAN~\cite{schonfeld2020u} & 7.48  & - & -\\
        & VQGAN~\cite{esser2021taming} & 11.40  & - & - \\
        & INR-GAN~\cite{skorokhodov2021adversarial} & 9.57 & - & -\\
        & CIPS~\cite{anokhin2021image} & 4.38 & - & -\\ \cmidrule{2-5}
        & StyleGAN2~\cite{karras2020analyzing} & 3.83 & 0.711 & 0.422 \\
        & IDM (Ours) & \textbf{3.54} & \textbf{0.720} & \textbf{0.425} \\         
     \bottomrule
    \end{tabular}
    }
    \caption{Comparison with the state-of-the-art methods on FFHQ in two resolutions $512\times512$ and $256\times256$.}
    \label{tab:ffhq_gen}
\end{table}

\begin{table}[!ht]
    \centering
    \resizebox{0.65\linewidth}{!}{
    \begin{tabular}{c|l|cc}
    \toprule
       Resolution & Methods & FID$\downarrow$ \\ \midrule
        \multirow{4}{*}{$\times 128$} & BigGAN~\cite{brock2018large} & 9.77\\
        & IDM (Ours) & \textbf{8.97}  \\ \cmidrule(r){2-3}
        & DDM & 14.43 \\
        & IDM (Ours) & \textbf{11.10} \\
     \bottomrule
    \end{tabular}
    }
    \caption[caption of imagenet gan]{Comparison with the BigGAN methods on conditional ImageNet $128\times128$ generation. For DDM-based experiment, we report the half-way results under the same number of iterations (100K).}
    \label{tab:imagenet_gen}
    \vspace{-6pt}
\end{table}

\subsection{Ablations}
As for ablations, we are mainly interested in how the setting of intrinsic and extrinsic iterations boost the restoration model's performance.

\paragraph{Intrinsic iteration}
We study the effect of the number of sampling steps and sampling schedule, \eg, DDPM~\cite{ho2020denoising} and DDIM~\cite{song2020denoising}. For the sampling schedule, DDIM yields inferior results such that we use DDPM everywhere in this paper.  Table~\ref{tab:num_steps} compares the performances by varying the number of diffusion steps. Increasing number of diffusion steps $>10$ doesn't give benefit so we choose 10 steps in this paper. It is equivalent to 2s to restore per image on V100 GPU. We also note that most single-forward pass methods take $<0.5$s. So the efficiency might be one limitation of the proposed approach.
\begin{table}[!ht]
    \centering
    \resizebox{0.85\linewidth}{!}{
    \begin{tabular}{l|ccccc}
    \toprule
        Steps & 5 & 10 & 15 & 20 \\ \midrule
        PSNR$\uparrow$ & 13.11 & 27.42 & 27.50 & 27.48 \\
        LPIPS$_\text{VGG}$ $\downarrow$ & 0.475 & 0.307 & 0.311 & 0.308 \\
    \bottomrule
    \end{tabular}
    }
    \caption{Varying the number of sampling steps.}
    \label{tab:num_steps}
    \vspace{-6pt}
\end{table}

\vspace{-4mm}
\paragraph{Extrinsic iteration}
We have claimed that extrinsic iteration enhances training data and improves restoration with another iteration of training. An example of the advantage of using extrinsic iterative learning is given below. Obviously, extrinsic learning helps increase clearness and reduce artifacts. An user study of evaluating this strategy also suggests that results from extrinsic learning are preferred.

\begin{figure}[!ht]
    \centering
    \setlength{\tabcolsep}{1.5pt}
    \renewcommand{\arraystretch}{0.8}
    \scalebox{0.8}{
    \begin{tabular}{c|cc}
        \includegraphics[width=0.32\linewidth]{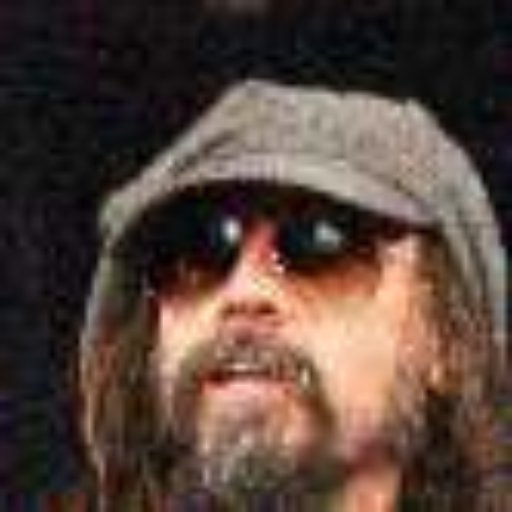} & 
        \includegraphics[width=0.32\linewidth]{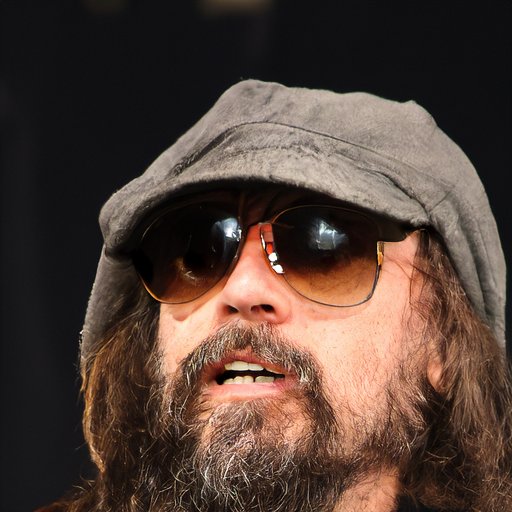} &
        \includegraphics[width=0.32\linewidth]{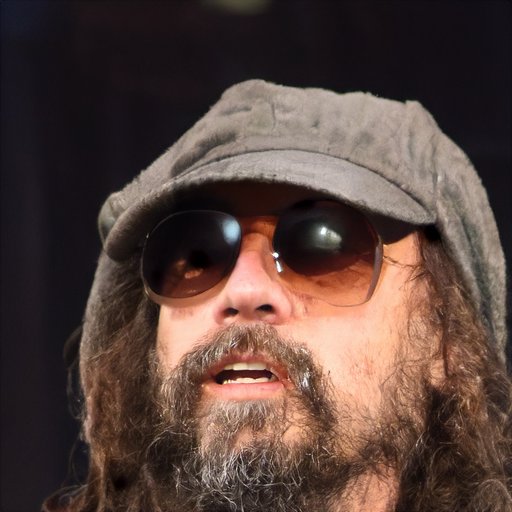}\\
        {\footnotesize Source} & {\footnotesize \textit{w/} extrinsic} & {\footnotesize \textit{w/o} extrinsic} \\
    \end{tabular}
    }
    \label{fig:extrinsic_merit}
    \vspace{-6pt}
\end{figure}

\vspace*{-0.05in}
\section{Conclusion}
\vspace*{-0.05in}
\label{sec:conclusion}

Face restoration has been an ill-posed problem. The importance of training data, testing data and evaluation criterion have been neglected. In this paper, we advocate an authentic restoration system with iteratively learned denoising diffusion models. The intrinsic iterative refinement mechanism can automatically maintain the high-quality details while erasing degradations. With the refined training data by such a learned model, we find it not only benefits restoration with another extrinsic iteration but also helps stabilize generation. Extensive experiment results have validated the effectiveness and authenticity of the system.

{\small
\bibliographystyle{ieee_fullname}
\bibliography{egbib}
}

\clearpage
\newpage
\appendix

\section{Implementation Details}

\paragraph{Training}
We set the global batch size to 32. If the loss is calculated on data $x$ directly, we run the training for 0.1M iterations. Otherwise, 1M iterations will be used following Eq.~\ref{eq:loss}. We use the Adam optimizer and apply a constant learning rate $1e^{-4}$.
Most existing works train DDM to predict the noise in the input. In contrast, our model predict the output, \ie~$x_0$, directly in the first intrinsic iteration. Empirical results show that this leads to a much faster convergence ($\approx 90\%$ faster) without affecting the final output quality compared with predicting noise.

\paragraph{Model}
The model architecture follows DDPM\footnote{https://github.com/hojonathanho/diffusion}. Conditional signals are concatenated to the U-Net and attentions are added to 8 - 32 resolutions.

\begin{figure*}
    \centering
    \setlength{\tabcolsep}{1pt}
    \renewcommand{\arraystretch}{0.8}
    \begin{tabular}{c|c|c}
    \toprule
        Source & $p=1$ & $p=2$ \\ \midrule
        \includegraphics[width=0.3\linewidth]{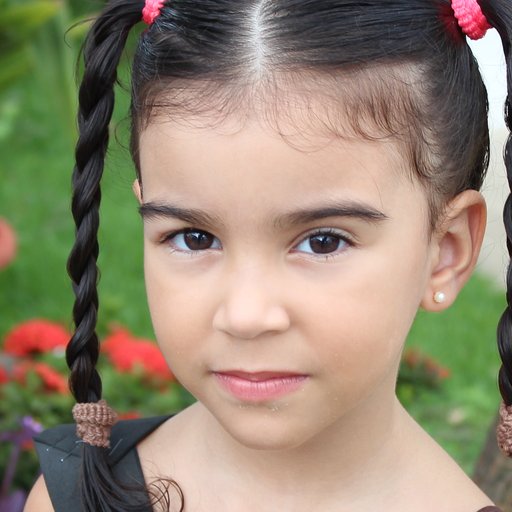} &
        \includegraphics[width=0.3\linewidth]{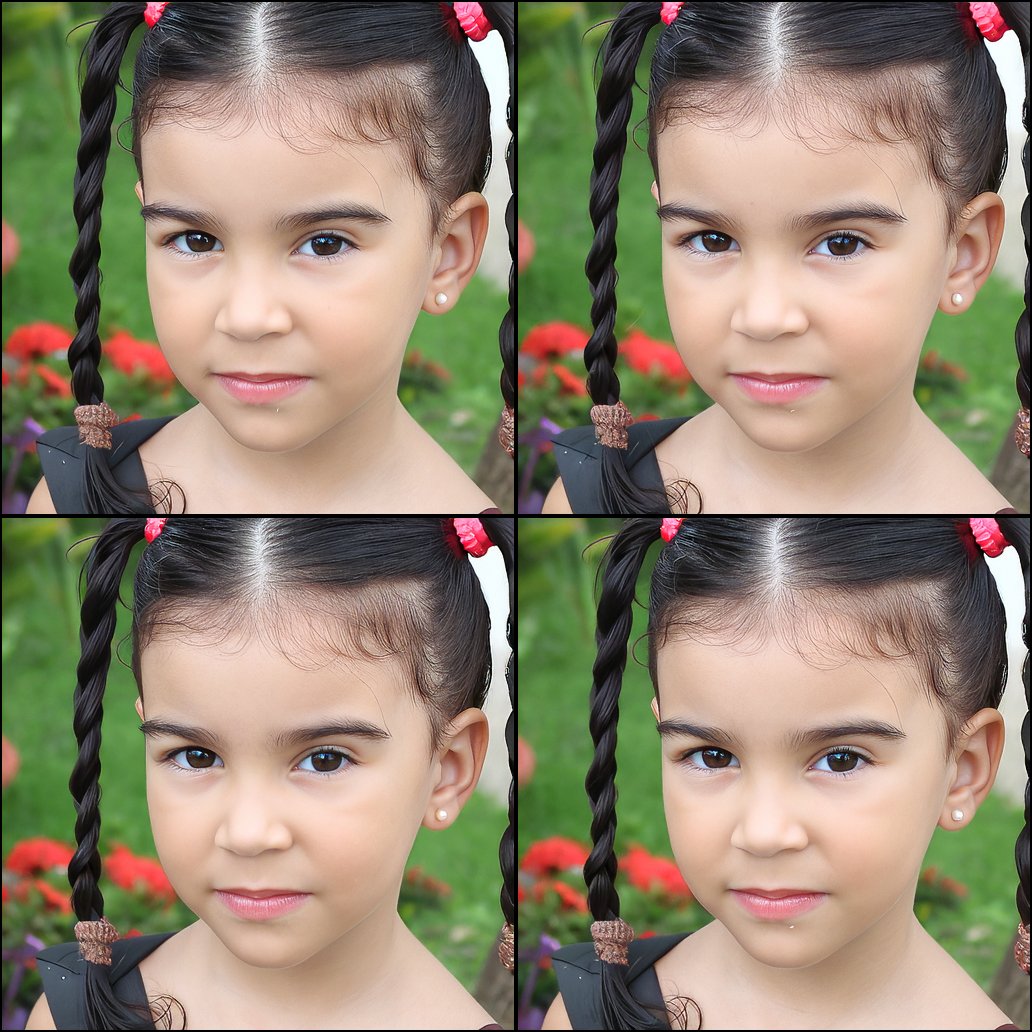} &
        \includegraphics[width=0.3\linewidth]{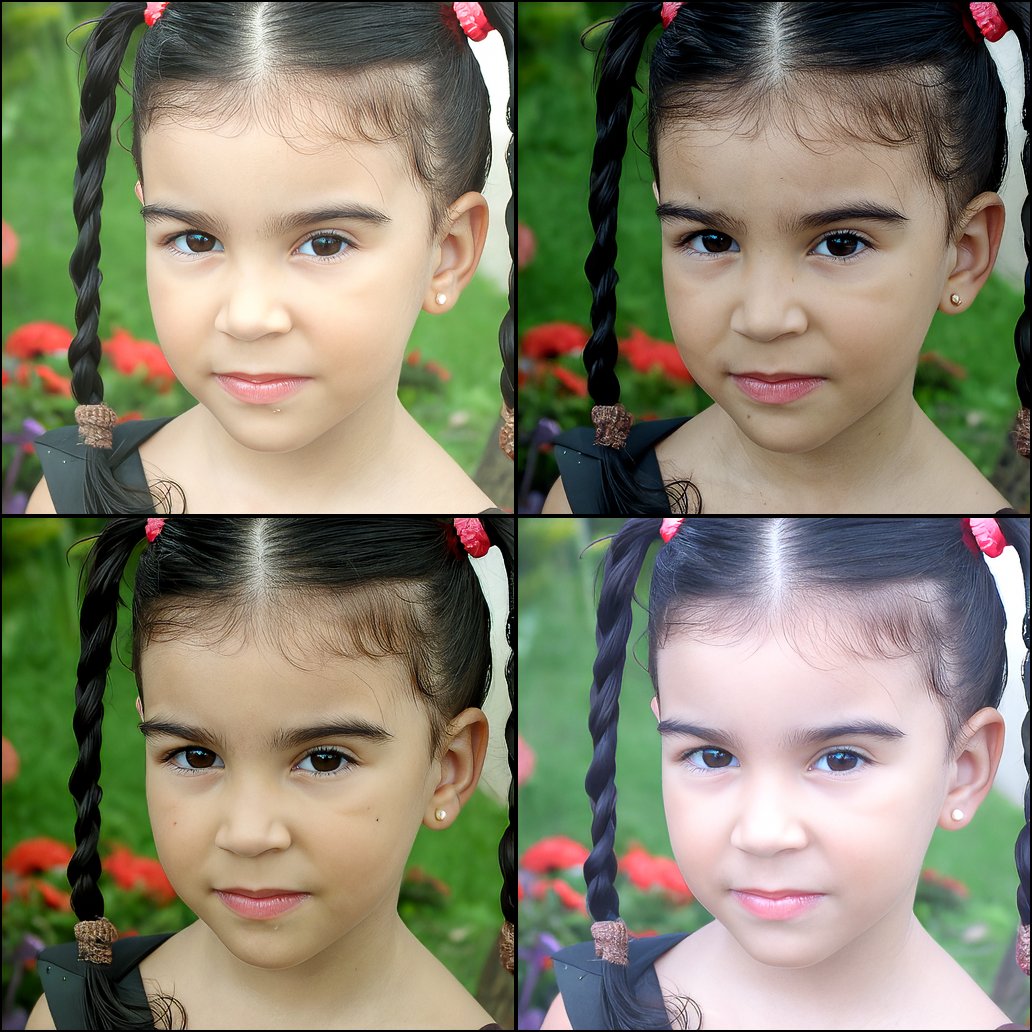} \\
        \includegraphics[width=0.3\linewidth]{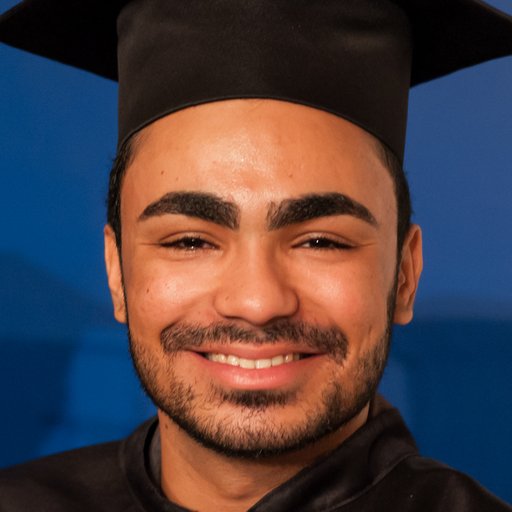} &
        \includegraphics[width=0.3\linewidth]{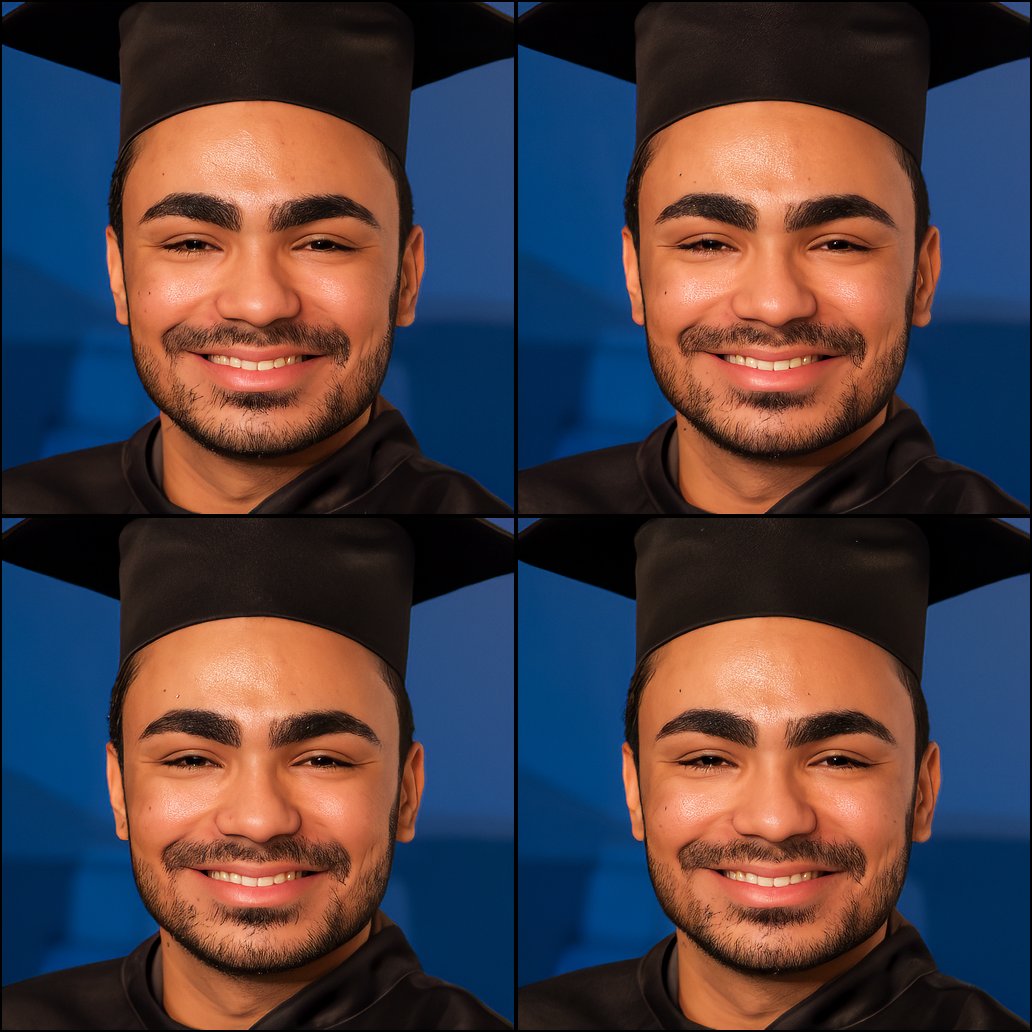} &
        \includegraphics[width=0.3\linewidth]{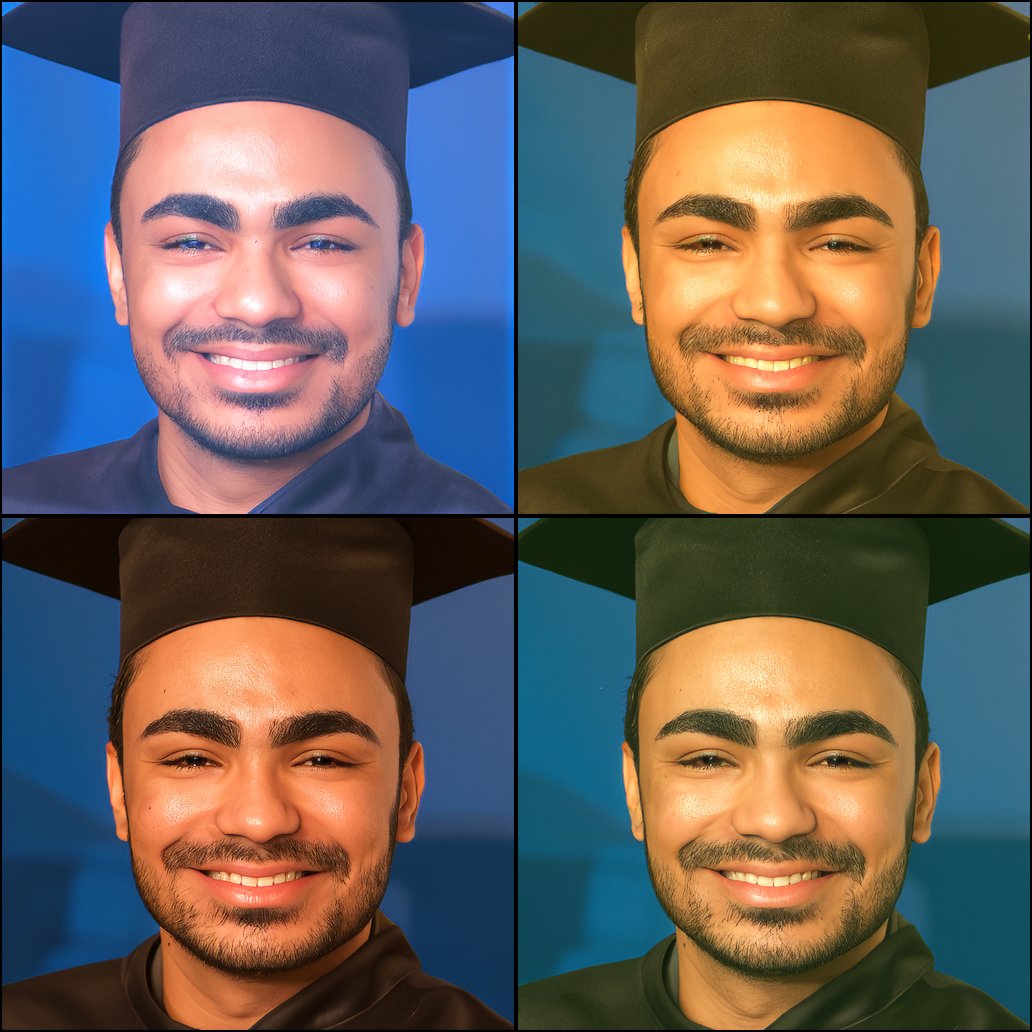} \\
        \includegraphics[width=0.3\linewidth]{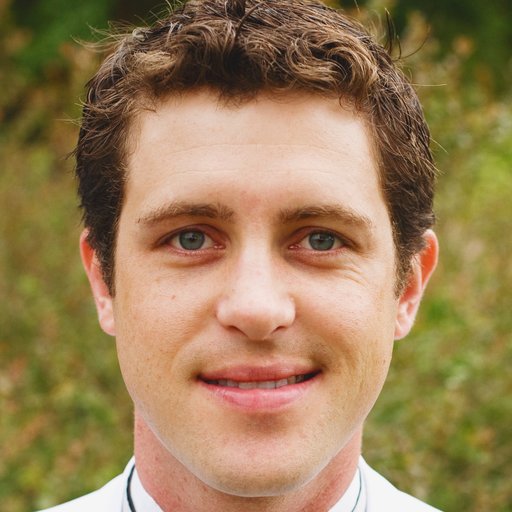} &
        \includegraphics[width=0.3\linewidth]{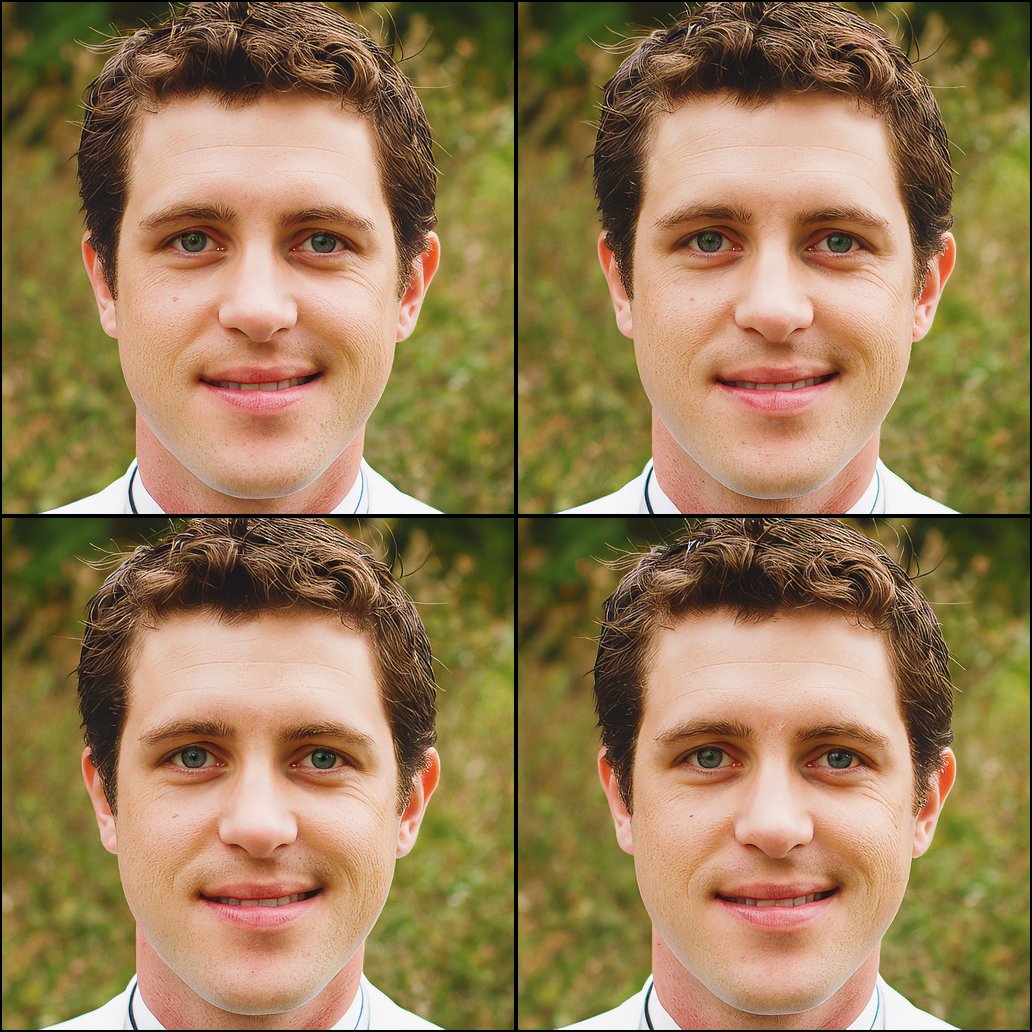} &
        \includegraphics[width=0.3\linewidth]{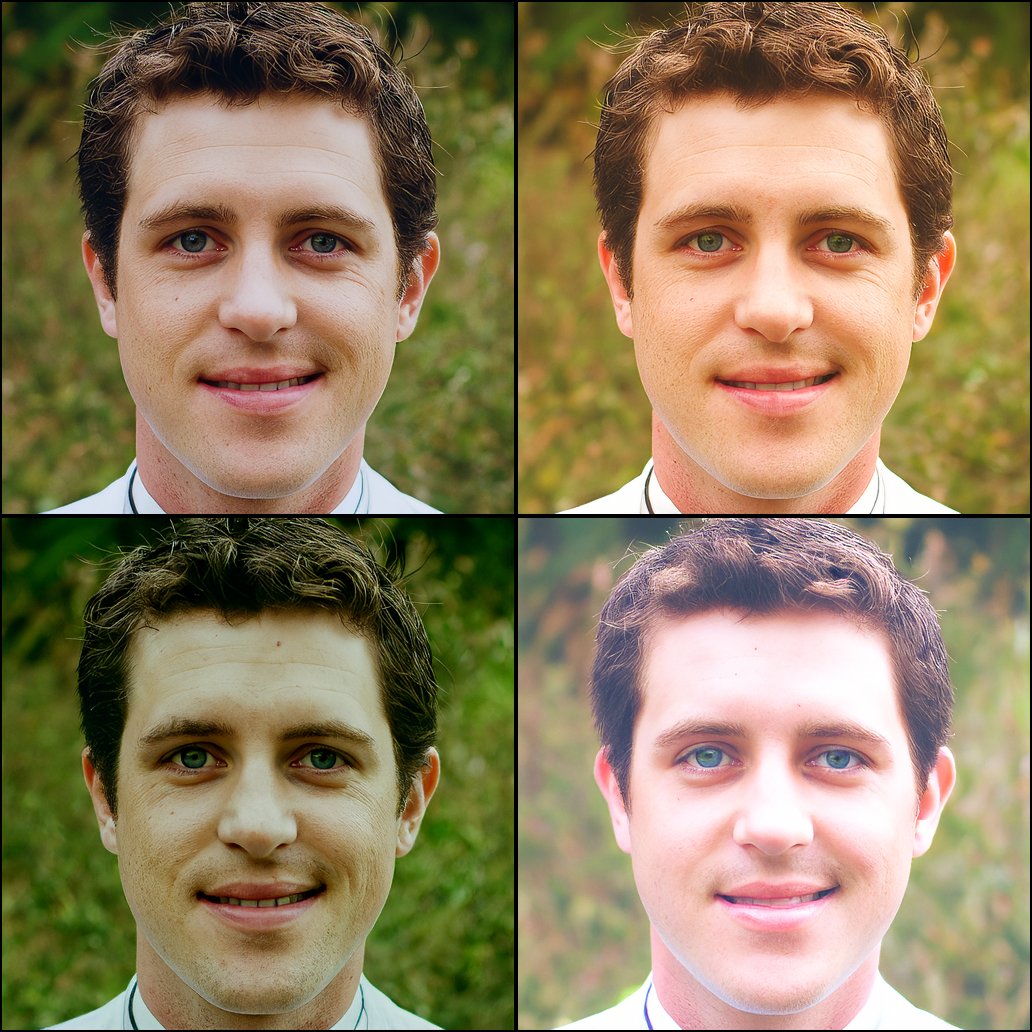} \\     
        \includegraphics[width=0.3\linewidth]{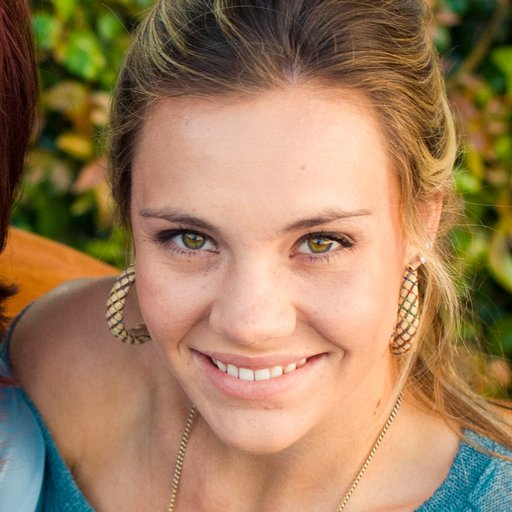} &
        \includegraphics[width=0.3\linewidth]{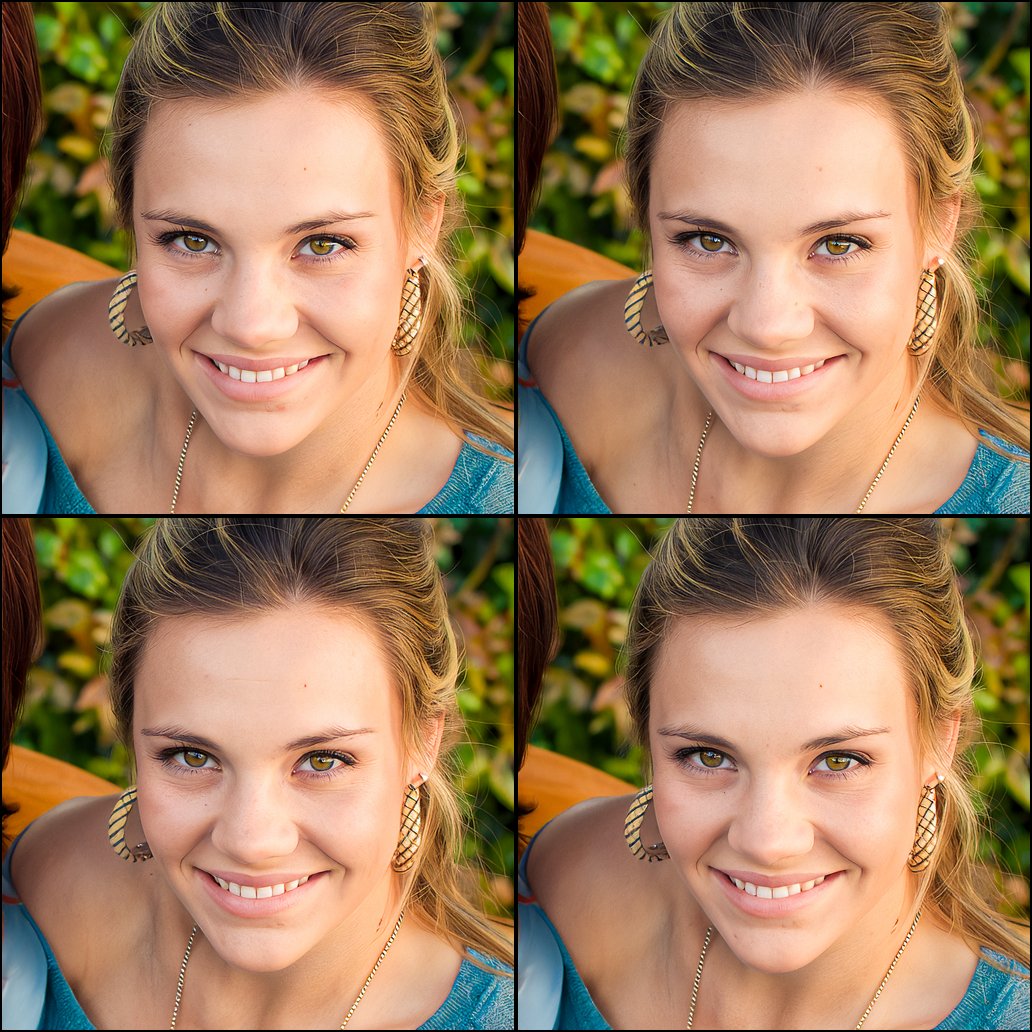} &
        \includegraphics[width=0.3\linewidth]{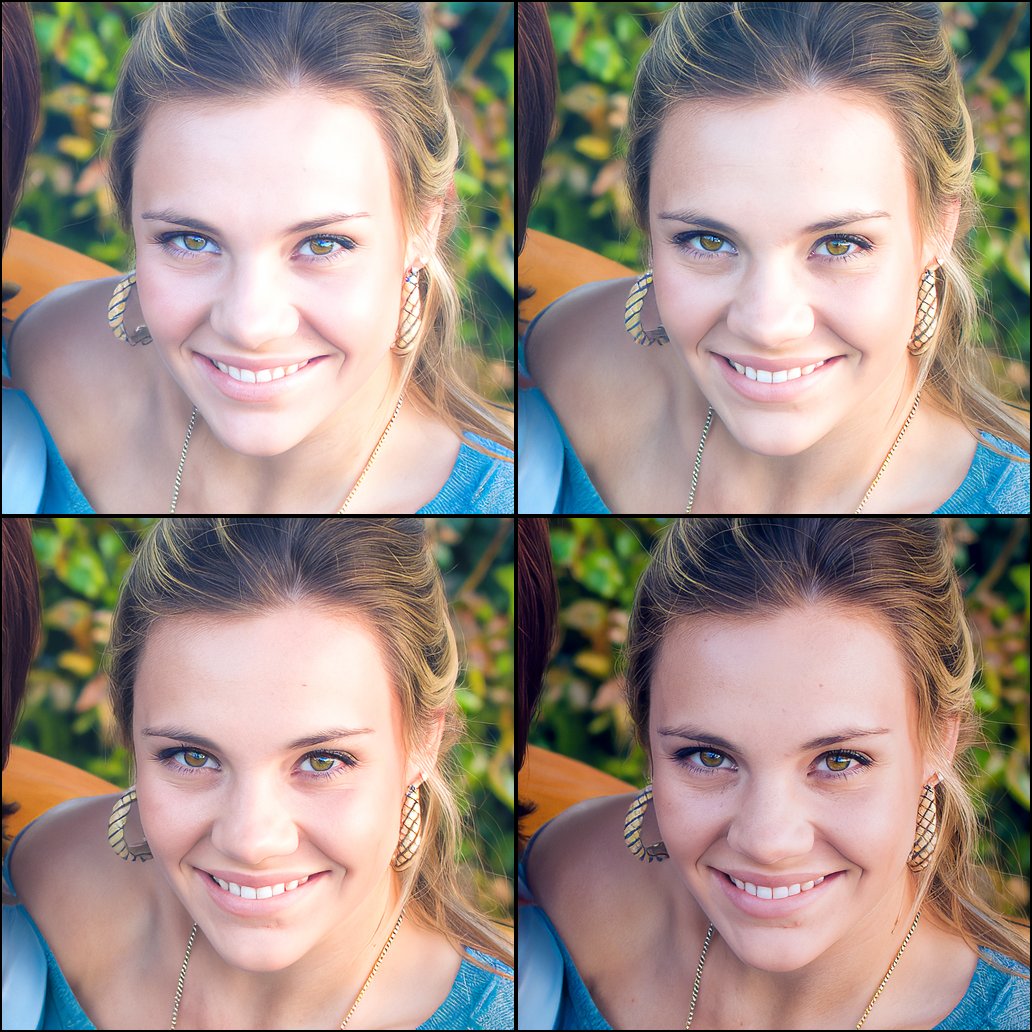} \\          
        \bottomrule
    \end{tabular}
    \caption{On the impact of $L_1$ and $L_2$. Examples come from the FFHQ dataset.}
    \label{fig:loss_norm_impact}
\end{figure*}

\paragraph{Loss}
In Eq.~\ref{eq:loss}, by choosing $p=1, 2$, we have some interesting observations. $p=1$ demonstrates much better stability in terms of color faithfulness. In contrast, $p=2$ yields obviously bad results, as shown in Figure~\ref{fig:loss_norm_impact}. For each input, we randomly sample four outputs. It could be due to optimizer settings and data fitting issues. Currently, we don't have a convincing explanation and we leave it for the future study.

\section{More examples}
\begin{figure*}[!h]
    \centering
    \resizebox{0.98\linewidth}{!}{
    \setlength{\tabcolsep}{1pt}
    \renewcommand{\arraystretch}{0.8}
    \begin{tabular}{c|cccc}
    \toprule
        Source & GFPGAN~\cite{wang2021towards} & CodeFormer~\cite{zhou2022towards} & Ours & GT \\ \midrule
        \includegraphics[width=0.18\linewidth]{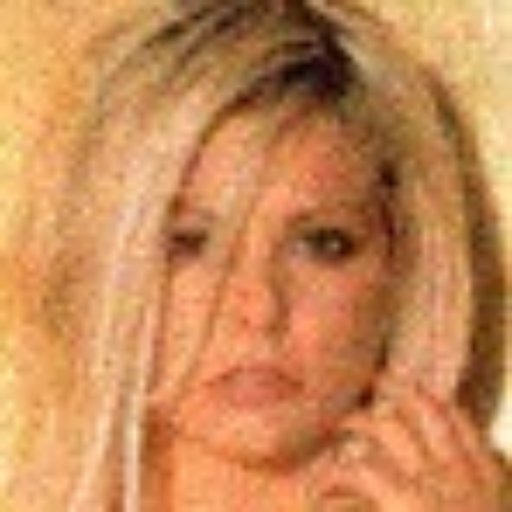} &
        \includegraphics[width=0.18\linewidth]{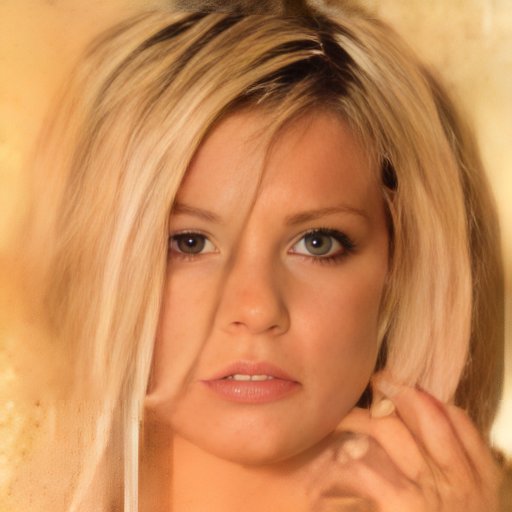} &
        \includegraphics[width=0.18\linewidth]{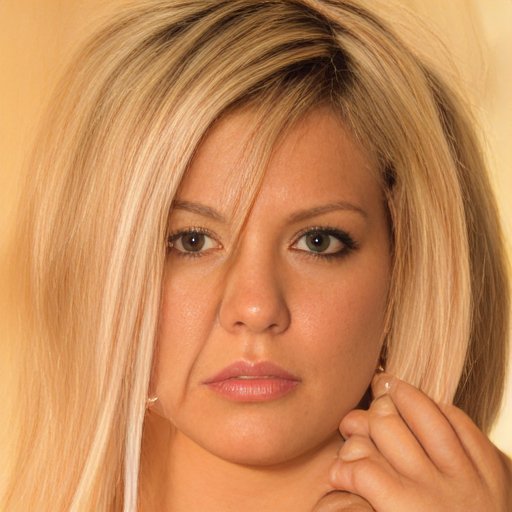}&
        \includegraphics[width=0.18\linewidth]{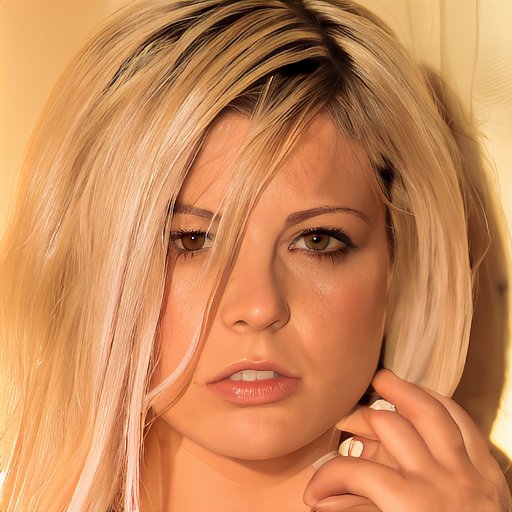}  & \includegraphics[width=0.18\linewidth]{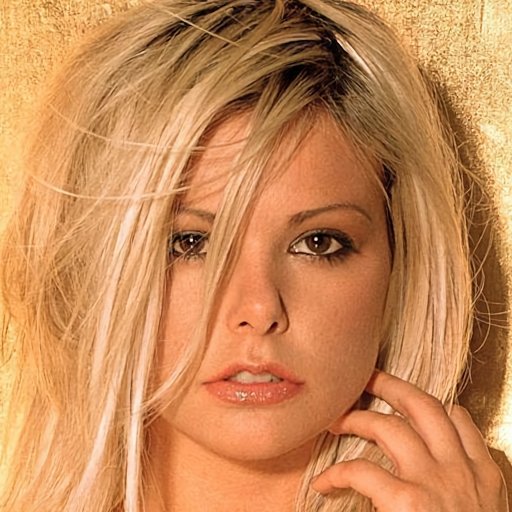} \\
        \includegraphics[width=0.18\linewidth]{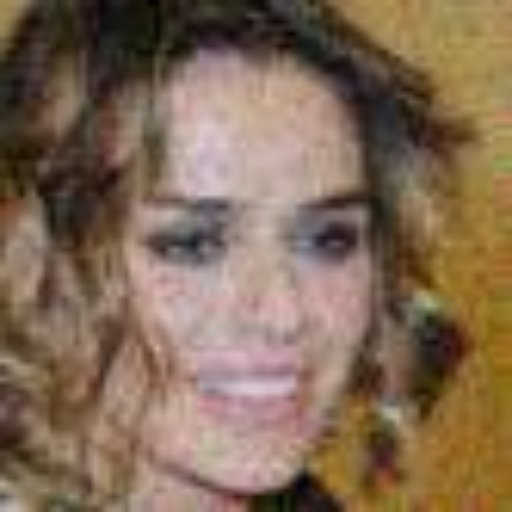} &
        \includegraphics[width=0.18\linewidth]{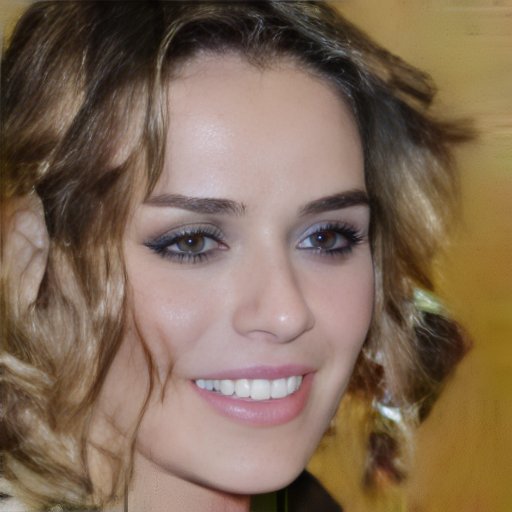}& 
        \includegraphics[width=0.18\linewidth]{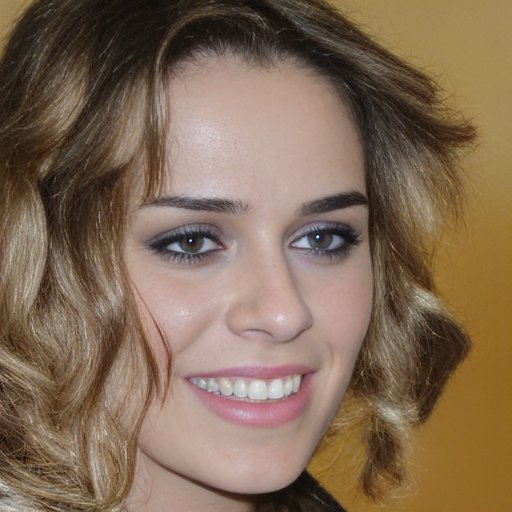} &
        \includegraphics[width=0.18\linewidth]{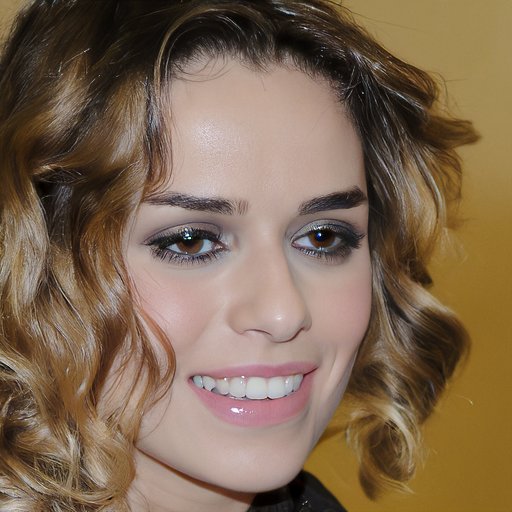} &  \includegraphics[width=0.18\linewidth]{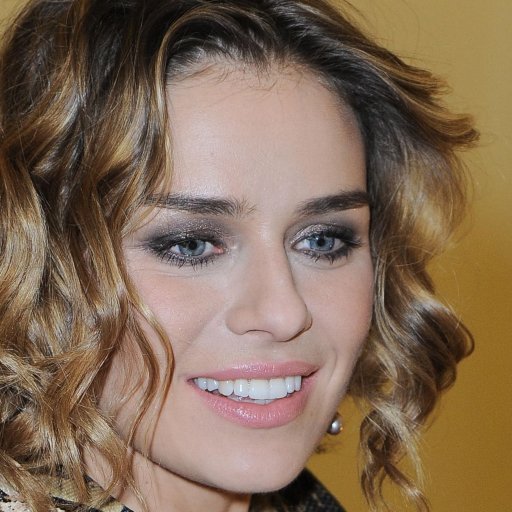}  \\ 
        \includegraphics[width=0.18\linewidth]{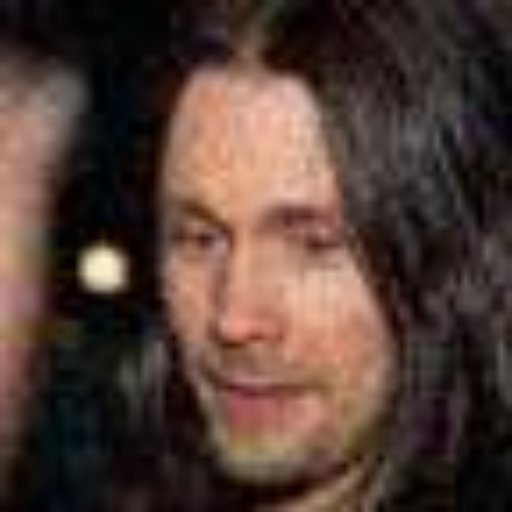} &
        \includegraphics[width=0.18\linewidth]{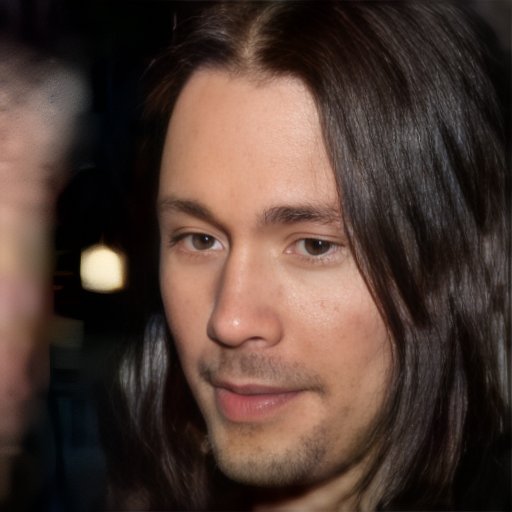}&
        \includegraphics[width=0.18\linewidth]{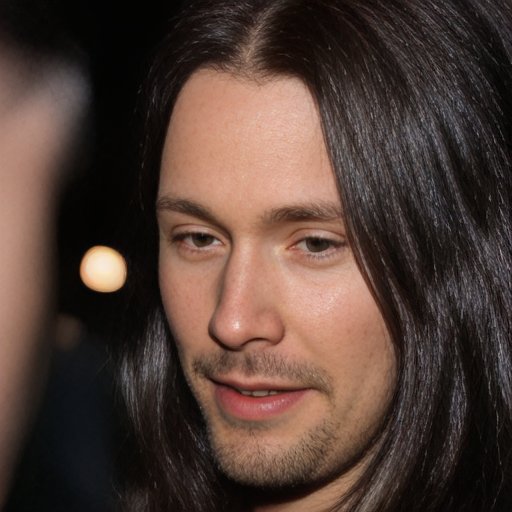} &
        \includegraphics[width=0.18\linewidth]{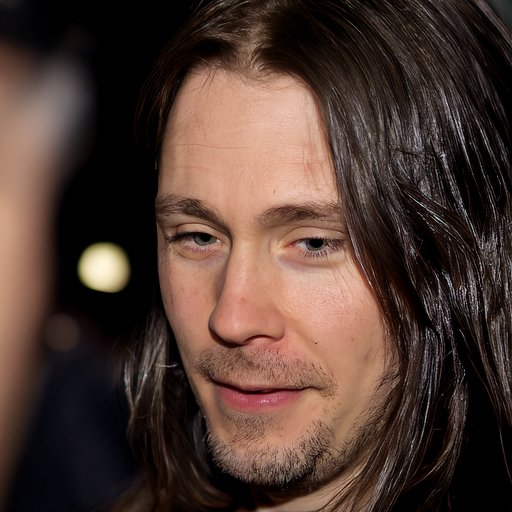}& \includegraphics[width=0.18\linewidth]{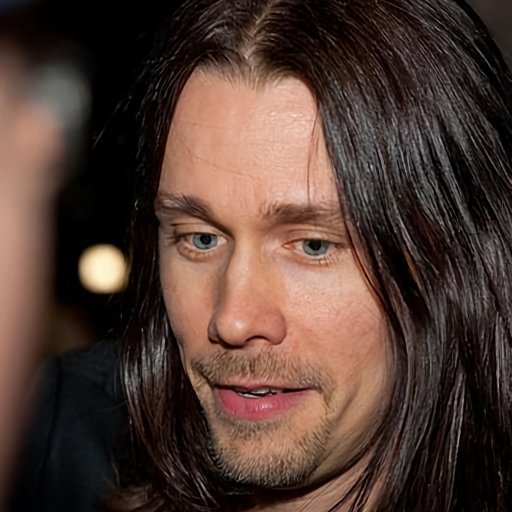} \\
        \includegraphics[width=0.18\linewidth]{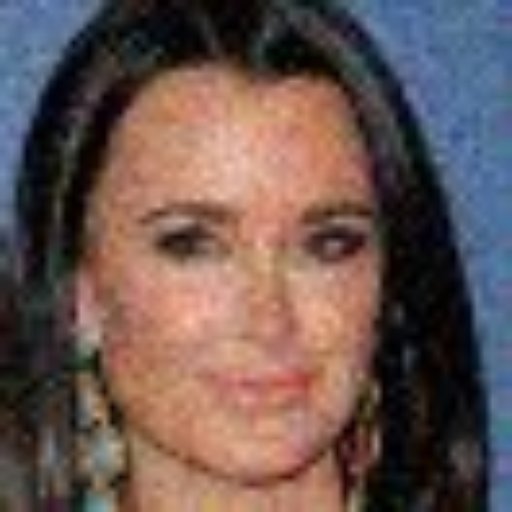} &
        \includegraphics[width=0.18\linewidth]{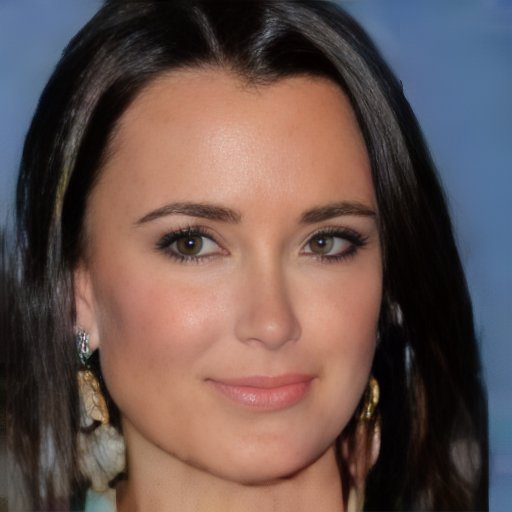}&
        \includegraphics[width=0.18\linewidth]{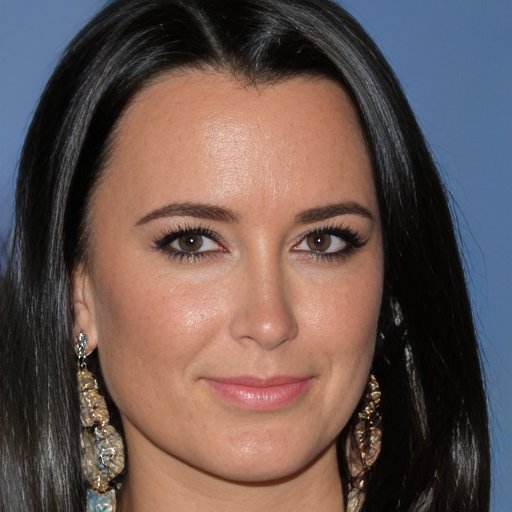} &
        \includegraphics[width=0.18\linewidth]{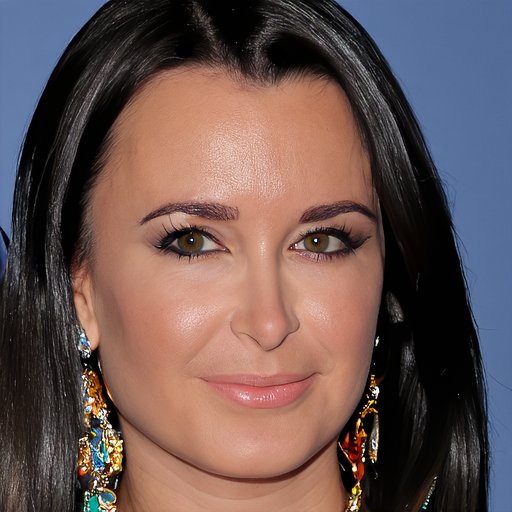}& \includegraphics[width=0.18\linewidth]{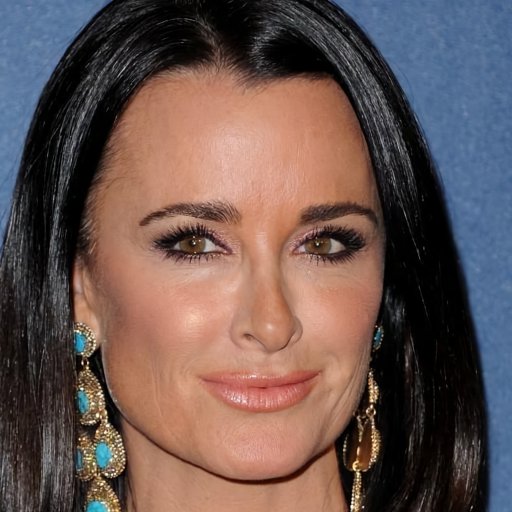} \\  
        \includegraphics[width=0.18\linewidth]{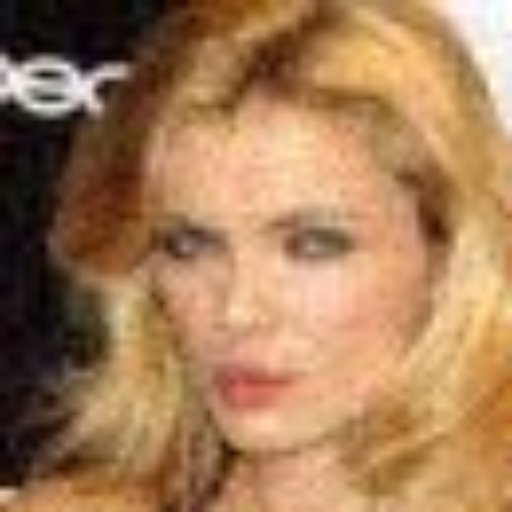} &
        \includegraphics[width=0.18\linewidth]{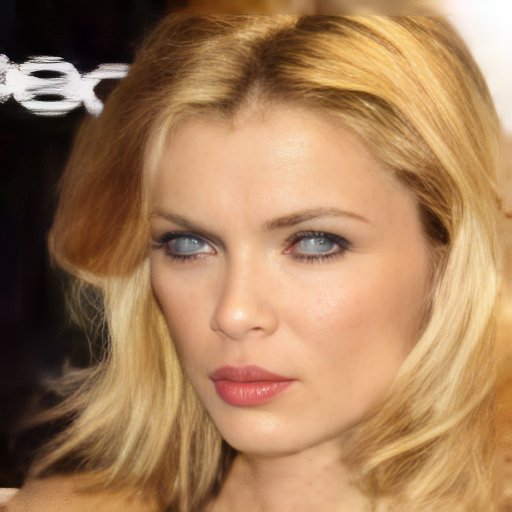}&
        \includegraphics[width=0.18\linewidth]{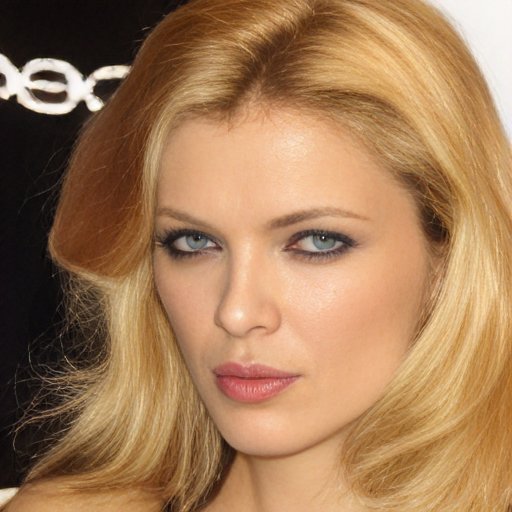} &
    \includegraphics[width=0.18\linewidth]{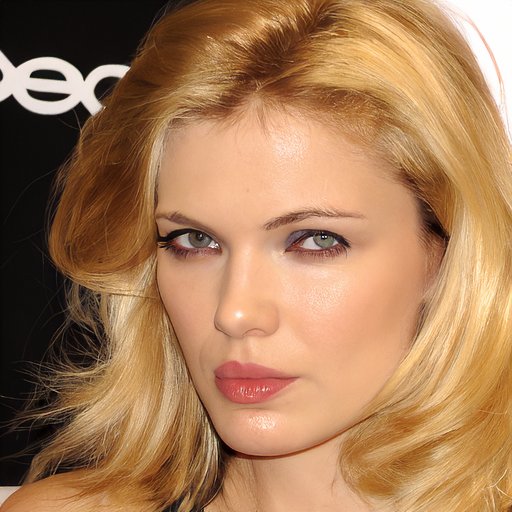}& \includegraphics[width=0.18\linewidth]{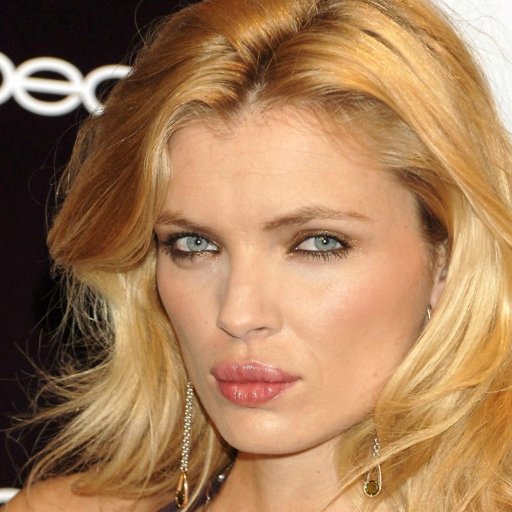} \\                 \includegraphics[width=0.18\linewidth]{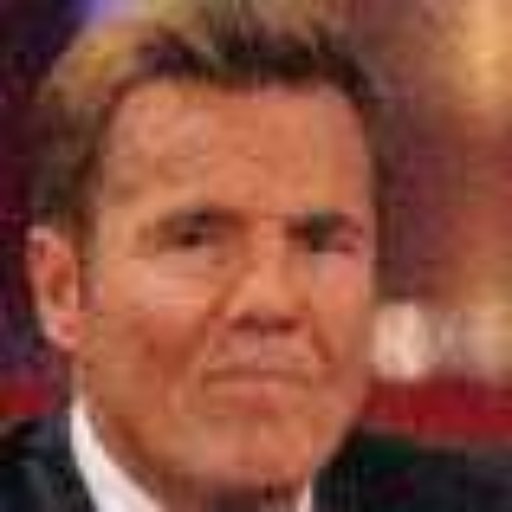} &
        \includegraphics[width=0.18\linewidth]{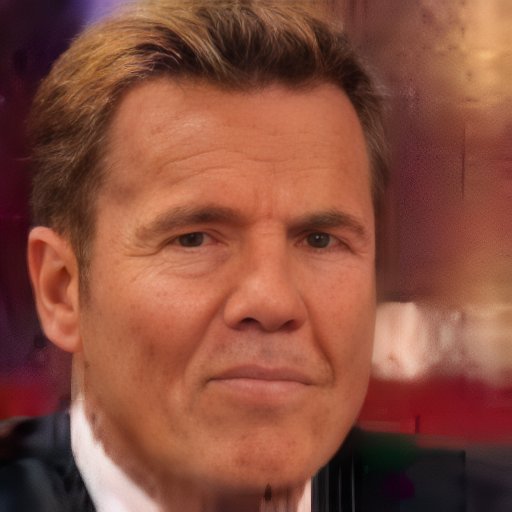}&
        \includegraphics[width=0.18\linewidth]{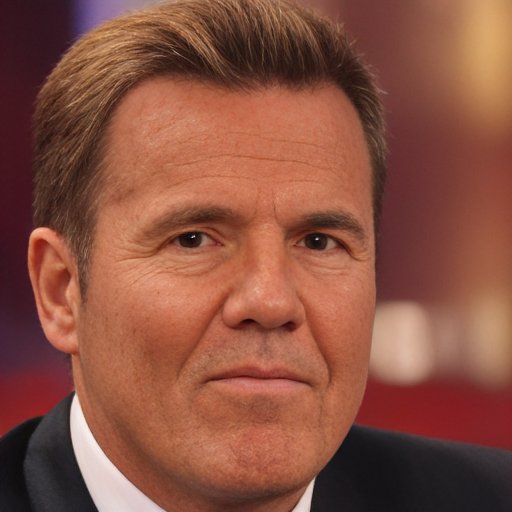} &
        \includegraphics[width=0.18\linewidth]{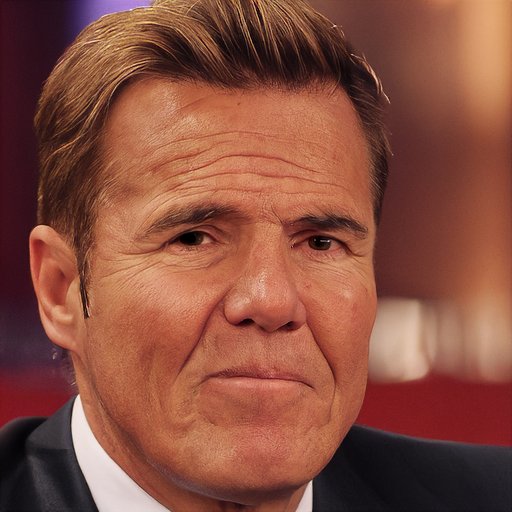}& \includegraphics[width=0.18\linewidth]{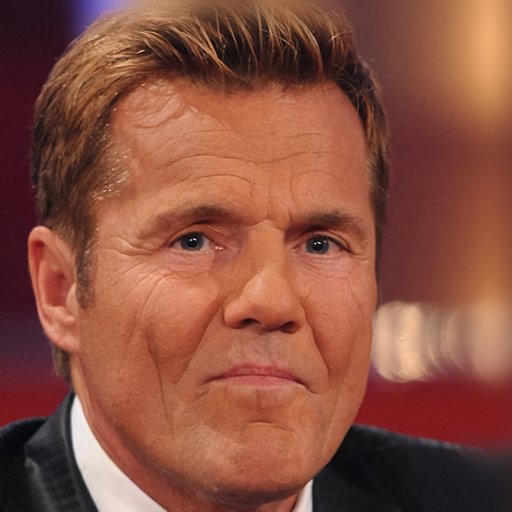} \\ 
    \bottomrule
    \end{tabular}
    }
    \caption{Qualitative comparison with state-of-the-art restoration models on synthetically degraded CelebA-HQ Test. }
    \label{fig:celebehq_restore_app}
\end{figure*}

\begin{figure*}[!h]
    \centering
    \resizebox{0.98\linewidth}{!}{
    \setlength{\tabcolsep}{1pt}
    \renewcommand{\arraystretch}{0.8}
    \begin{tabular}{c|ccc}
    \toprule
        Source & GFPGAN~\cite{wang2021towards} & CodeFormer~\cite{zhou2022towards} & Ours\\ \midrule
        \includegraphics[width=0.18\linewidth]{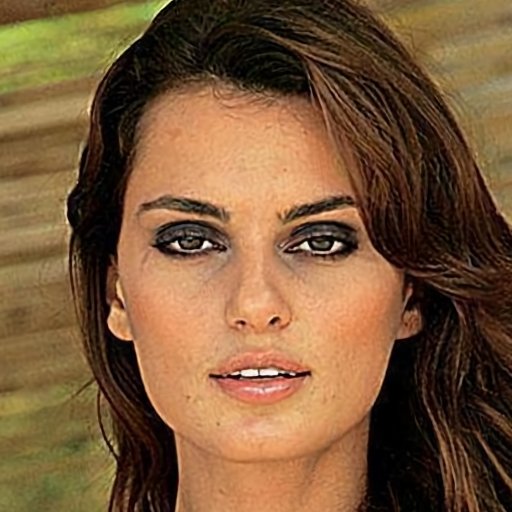} &
        \includegraphics[width=0.18\linewidth]{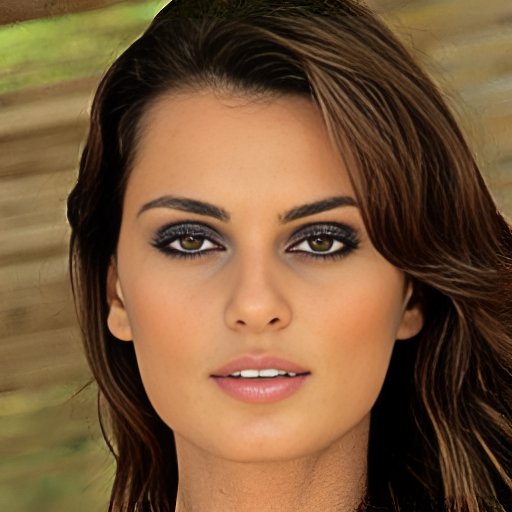} &
        \includegraphics[width=0.18\linewidth]{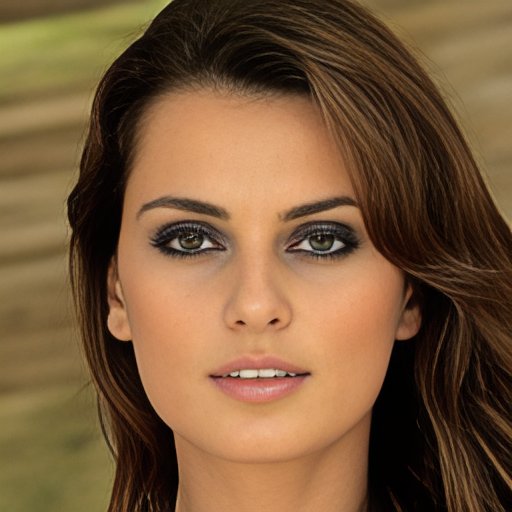}&
        \includegraphics[width=0.18\linewidth]{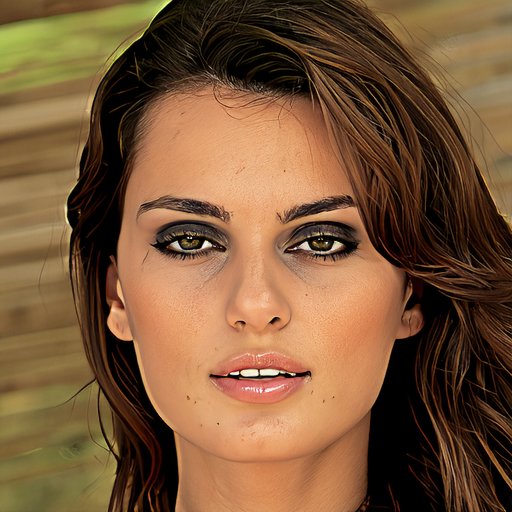}  \\
        \includegraphics[width=0.18\linewidth]{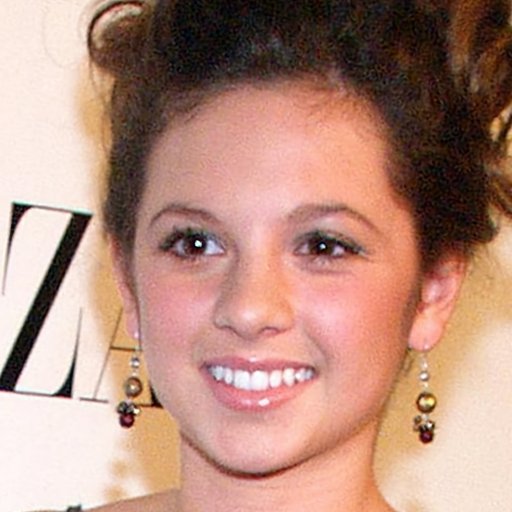} &
        \includegraphics[width=0.18\linewidth]{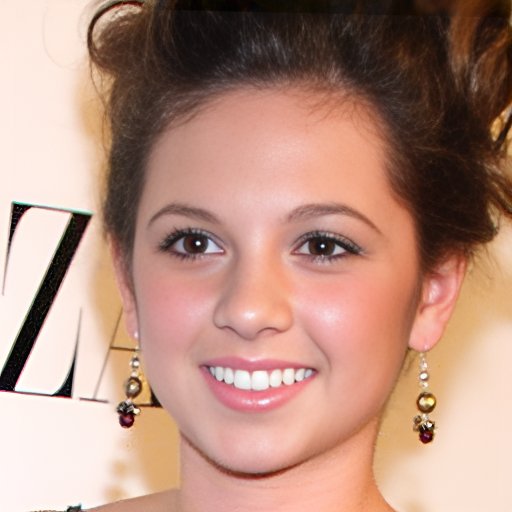}& 
        \includegraphics[width=0.18\linewidth]{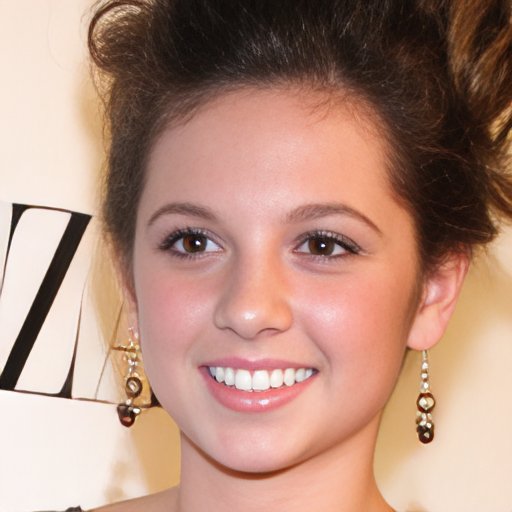}& 
        \includegraphics[width=0.18\linewidth]{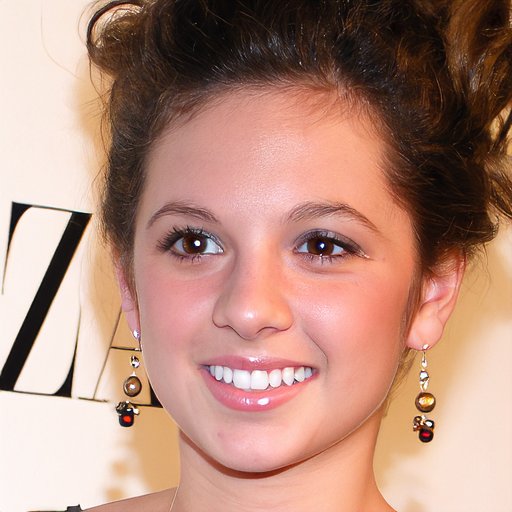}  \\ 
        \includegraphics[width=0.18\linewidth]{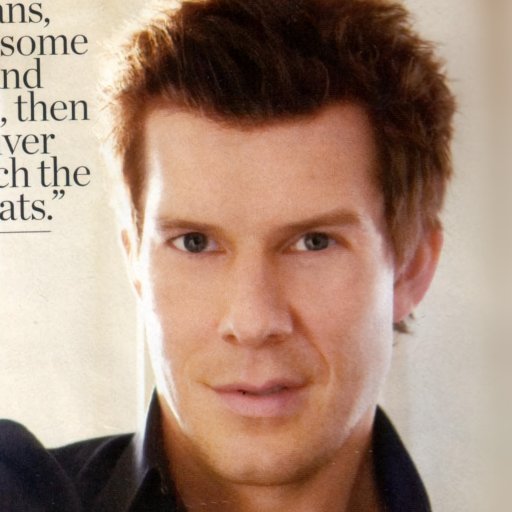} &
        \includegraphics[width=0.18\linewidth]{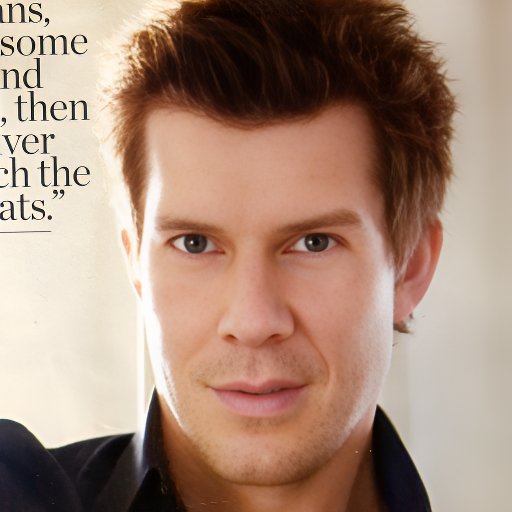}&
        \includegraphics[width=0.18\linewidth]{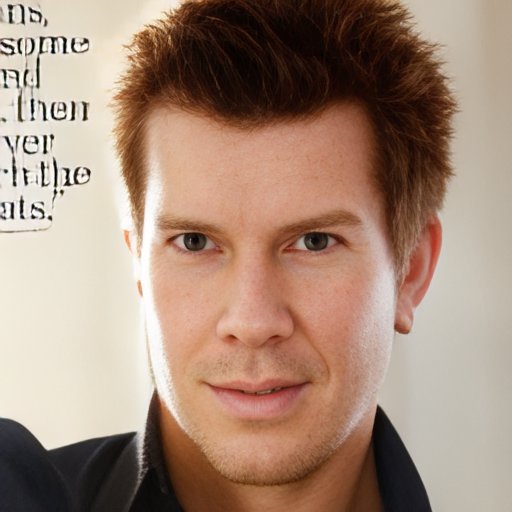} &
        \includegraphics[width=0.18\linewidth]{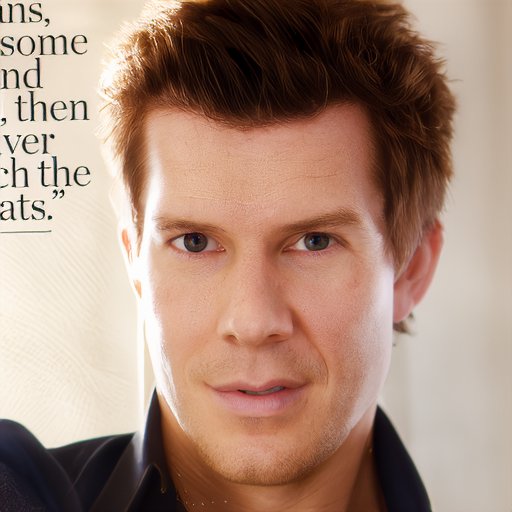} \\
        \includegraphics[width=0.18\linewidth]{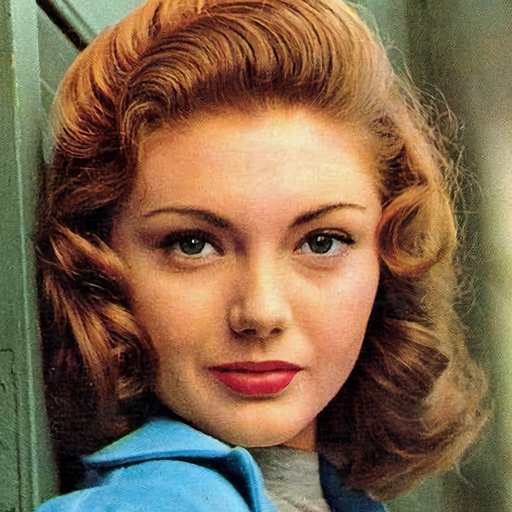} &
        \includegraphics[width=0.18\linewidth]{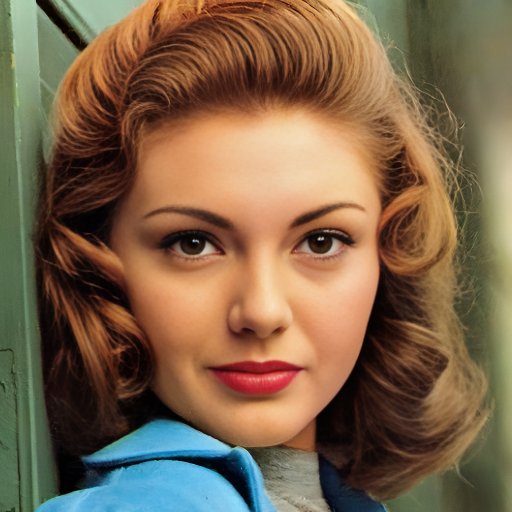}&
        \includegraphics[width=0.18\linewidth]{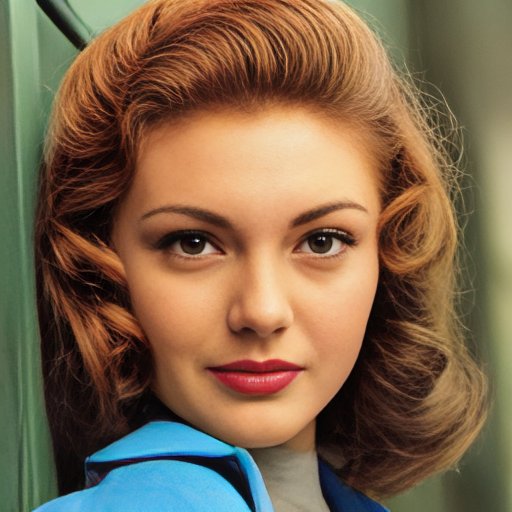} &
        \includegraphics[width=0.18\linewidth]{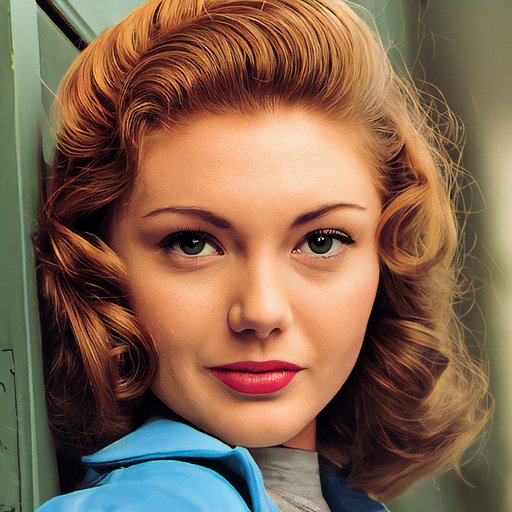} \\      
        \includegraphics[width=0.18\linewidth]{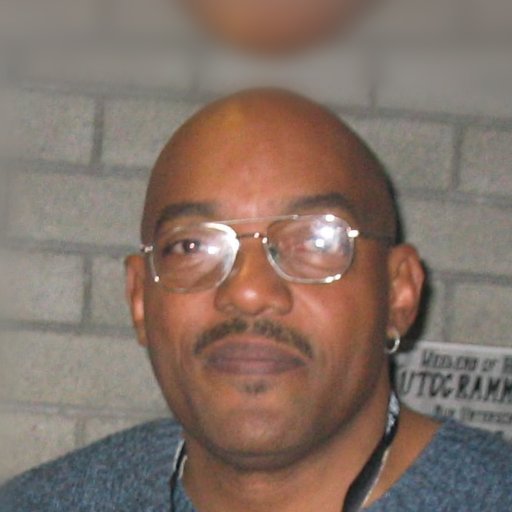} &
        \includegraphics[width=0.18\linewidth]{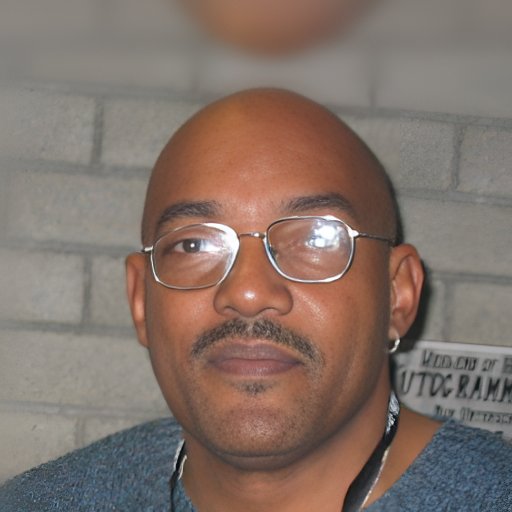}&
        \includegraphics[width=0.18\linewidth]{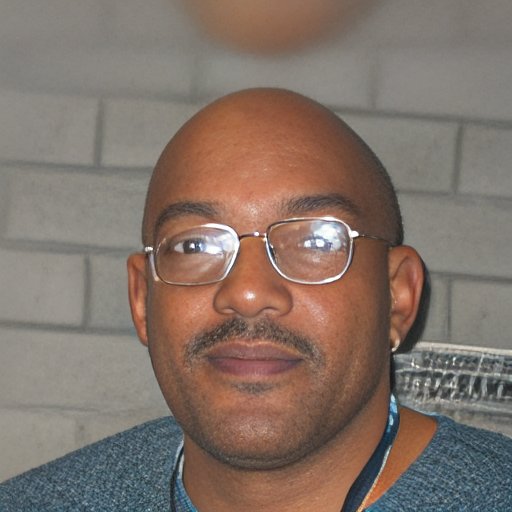} &
        \includegraphics[width=0.18\linewidth]{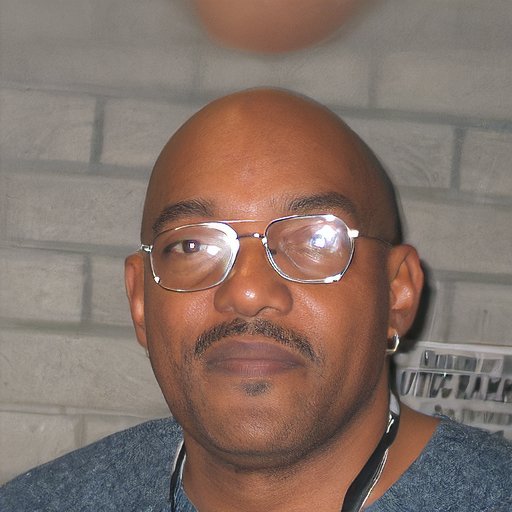} \\          
    \bottomrule
    \end{tabular}
    }
    \caption{Qualitative comparison with state-of-the-art restoration models on original CelebA-HQ Test. }
    \label{fig:celebehq_deg_restore_app}
\end{figure*}

\begin{figure*}[!h]
    \centering
    \resizebox{0.98\linewidth}{!}{
    \setlength{\tabcolsep}{1pt}
    \renewcommand{\arraystretch}{0.8}
    \begin{tabular}{c|ccc}
    \toprule
        Source & GFPGAN~\cite{wang2021towards} & CodeFormer~\cite{zhou2022towards} & Ours\\ \midrule
        \includegraphics[width=0.18\linewidth]{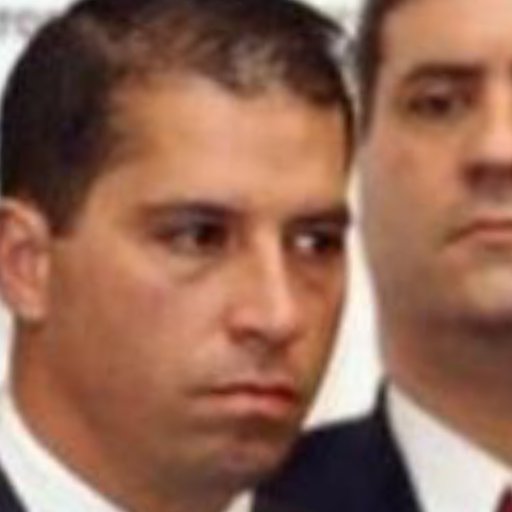} &
        \includegraphics[width=0.18\linewidth]{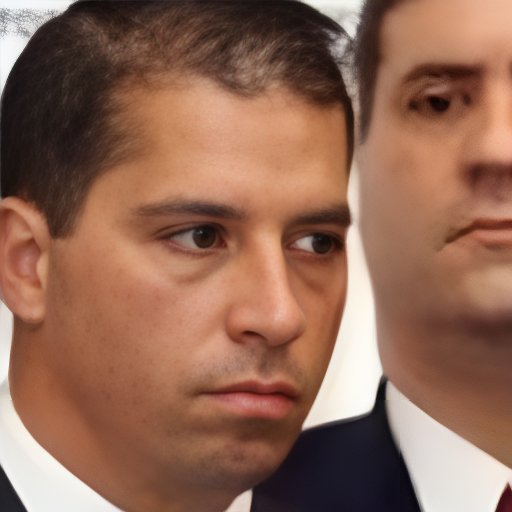} &
        \includegraphics[width=0.18\linewidth]{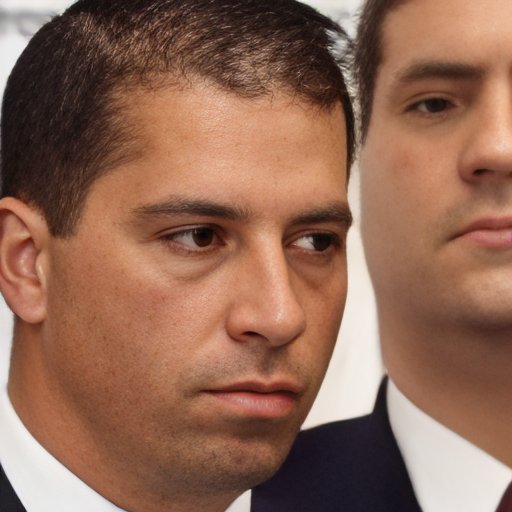}&
        \includegraphics[width=0.18\linewidth]{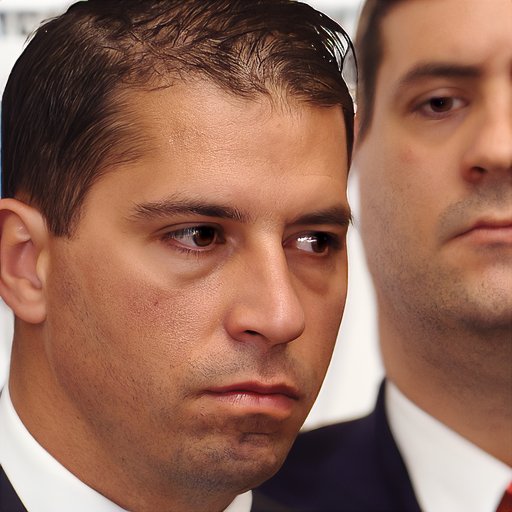}  \\
        \includegraphics[width=0.18\linewidth]{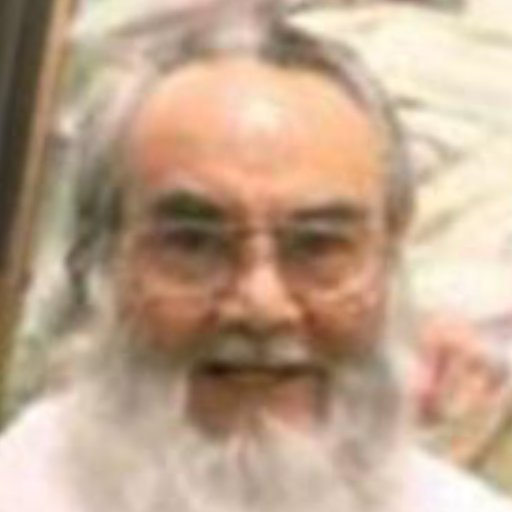} &
        \includegraphics[width=0.18\linewidth]{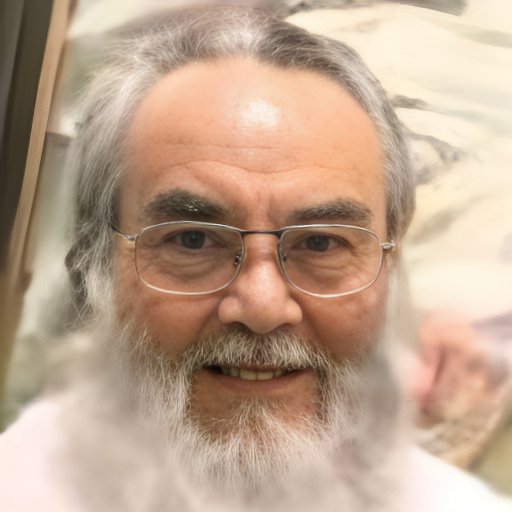}& 
        \includegraphics[width=0.18\linewidth]{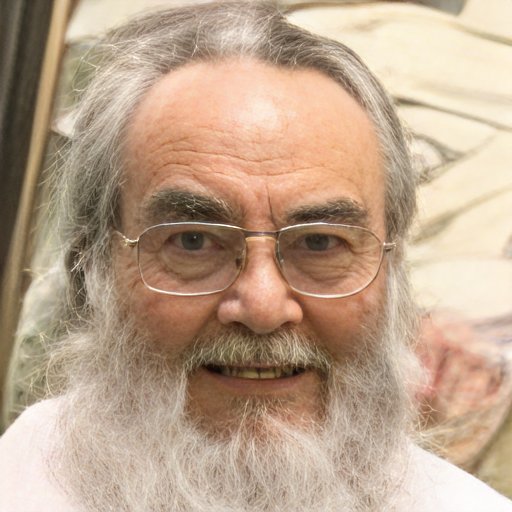}& 
        \includegraphics[width=0.18\linewidth]{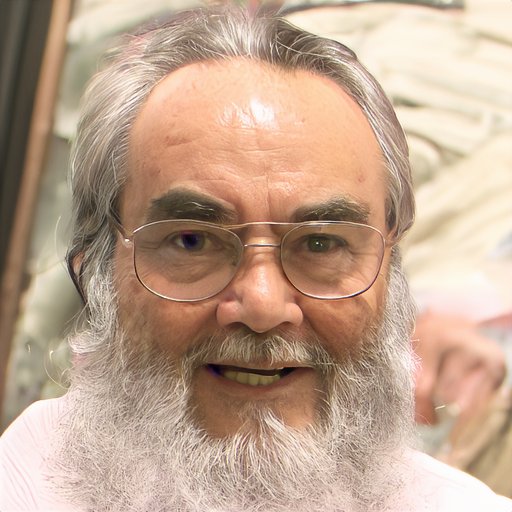}  \\ 
        \includegraphics[width=0.18\linewidth]{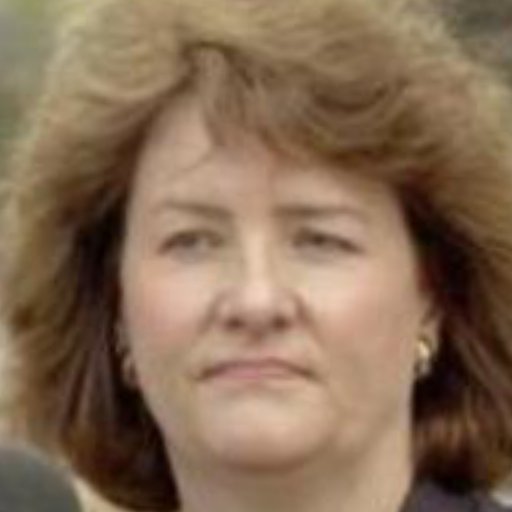} &
        \includegraphics[width=0.18\linewidth]{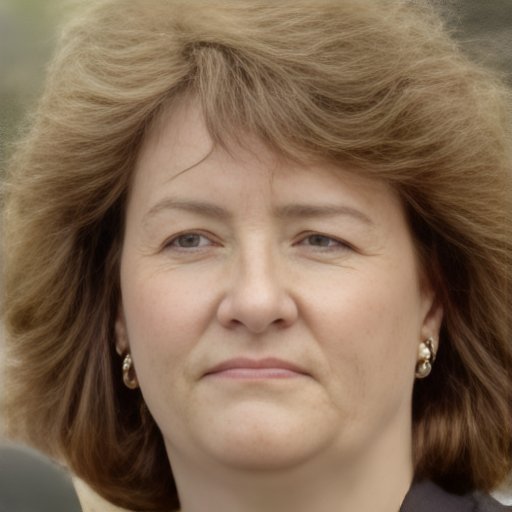}&
        \includegraphics[width=0.18\linewidth]{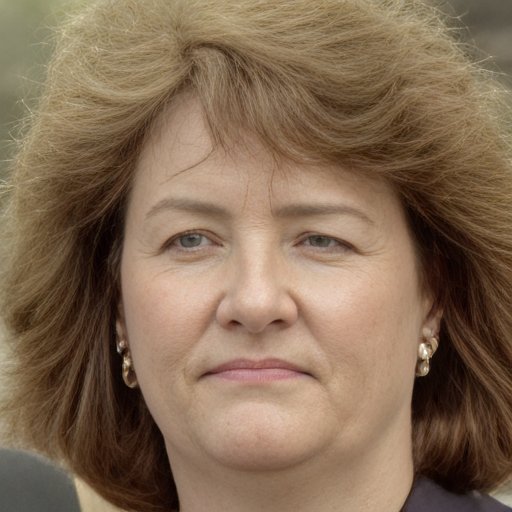} &
        \includegraphics[width=0.18\linewidth]{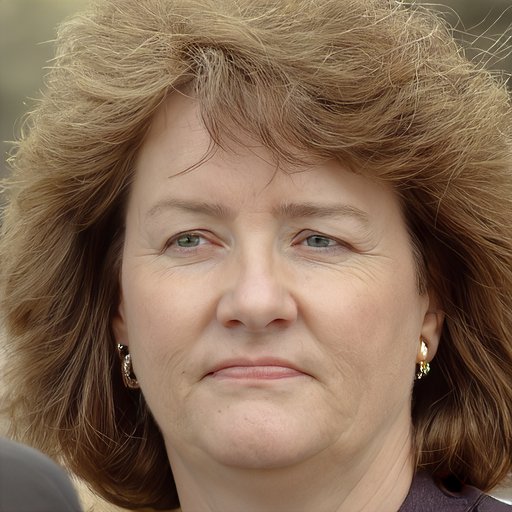} \\
        \includegraphics[width=0.18\linewidth]{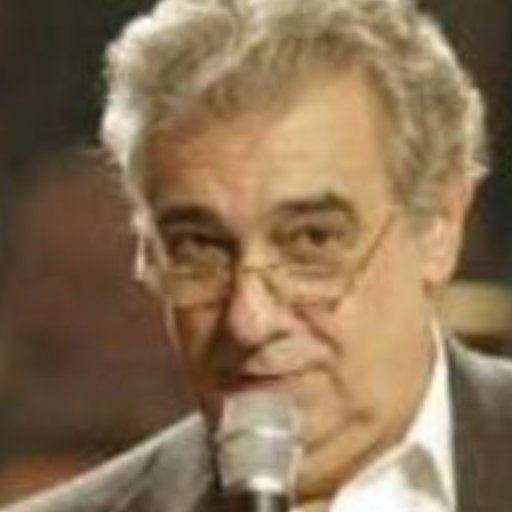} &
        \includegraphics[width=0.18\linewidth]{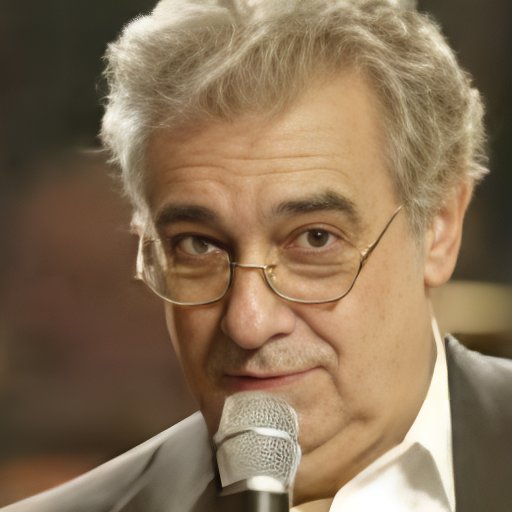}&
        \includegraphics[width=0.18\linewidth]{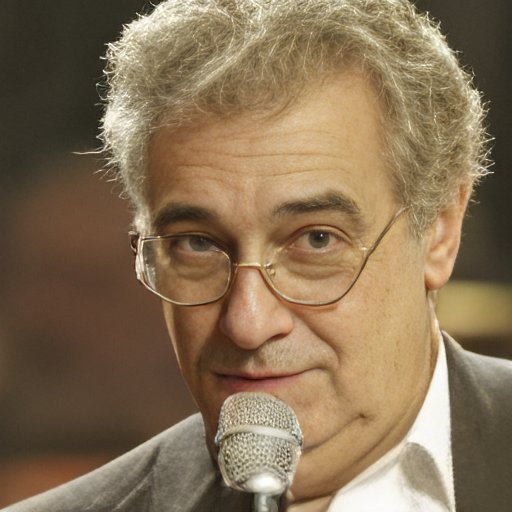} &
        \includegraphics[width=0.18\linewidth]{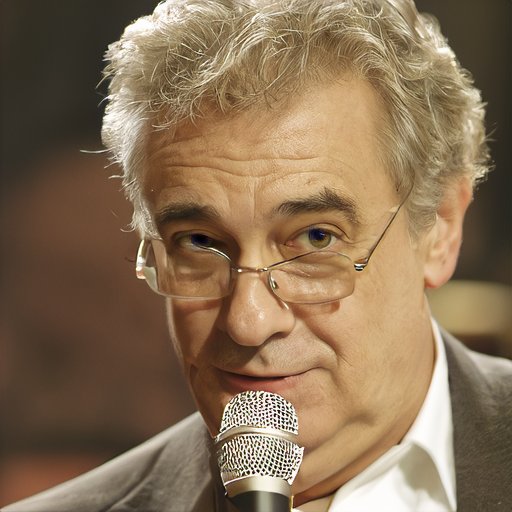} \\      
        \includegraphics[width=0.18\linewidth]{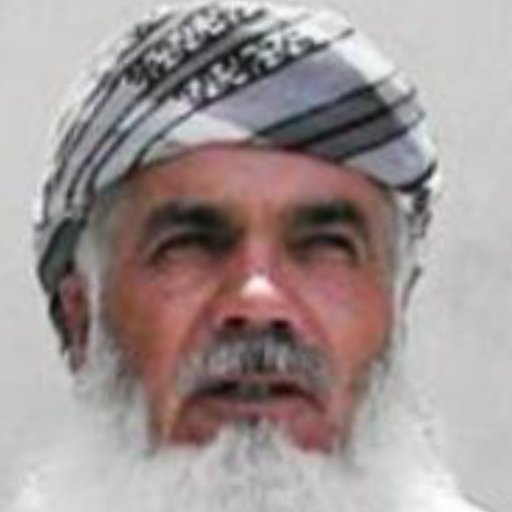} &
        \includegraphics[width=0.18\linewidth]{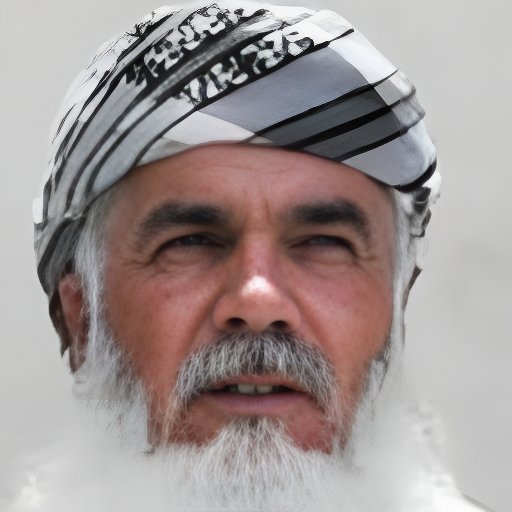}&
        \includegraphics[width=0.18\linewidth]{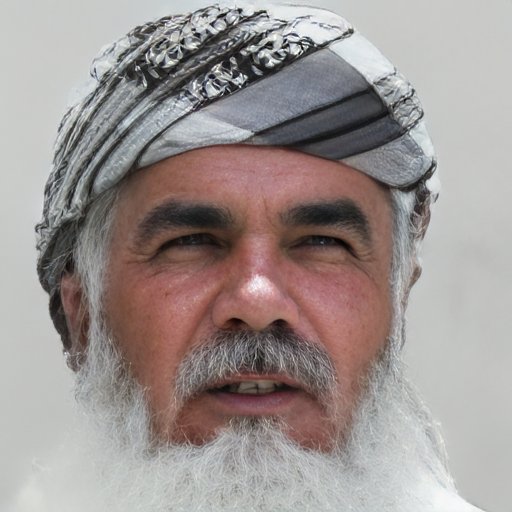} &
        \includegraphics[width=0.18\linewidth]{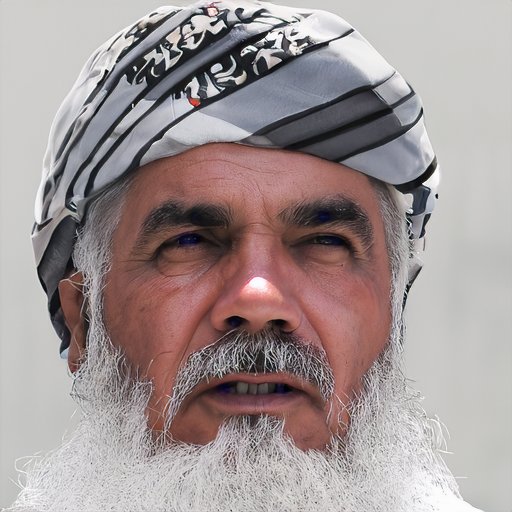} \\          
    \bottomrule
    \end{tabular}
    }
    \caption{Qualitative comparison with state-of-the-art restoration models on LFW test set.}
    \label{fig:lfw_app}
\end{figure*}

\begin{figure*}[!h]
    \centering
    \resizebox{0.98\linewidth}{!}{
    \setlength{\tabcolsep}{1pt}
    \renewcommand{\arraystretch}{0.8}
    \begin{tabular}{c|ccc}
    \toprule
        Source & GFPGAN~\cite{wang2021towards} & CodeFormer~\cite{zhou2022towards} & Ours\\ \midrule
        \includegraphics[width=0.18\linewidth]{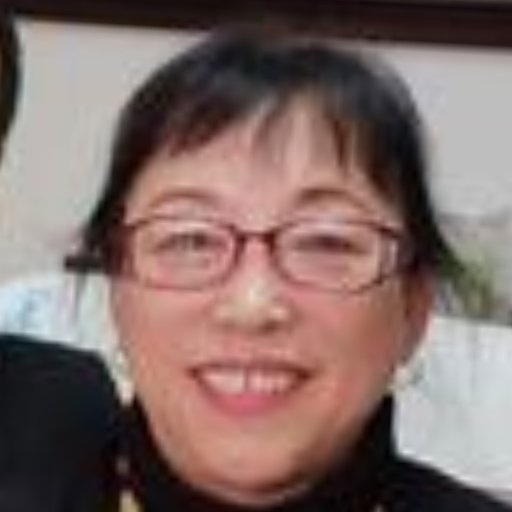} &
        \includegraphics[width=0.18\linewidth]{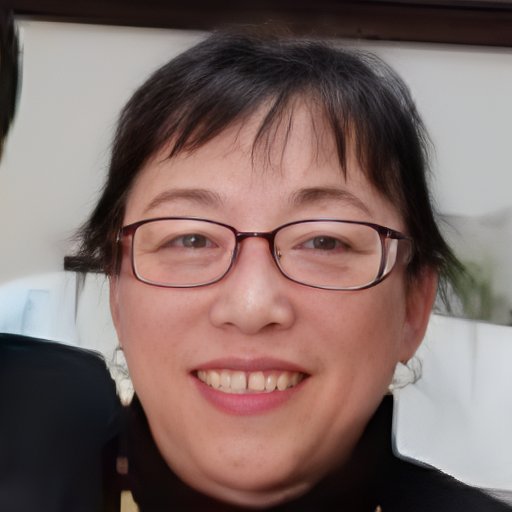} &
        \includegraphics[width=0.18\linewidth]{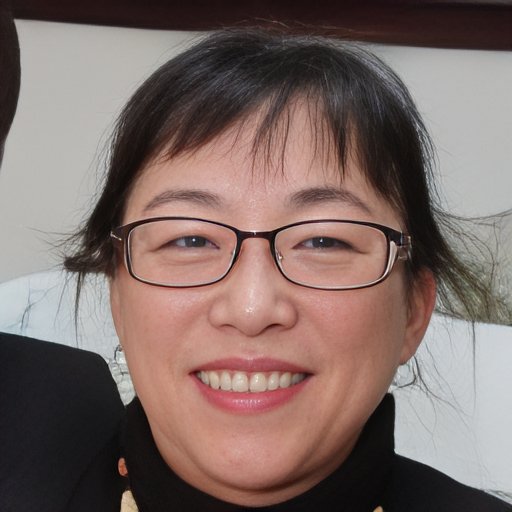}&
        \includegraphics[width=0.18\linewidth]{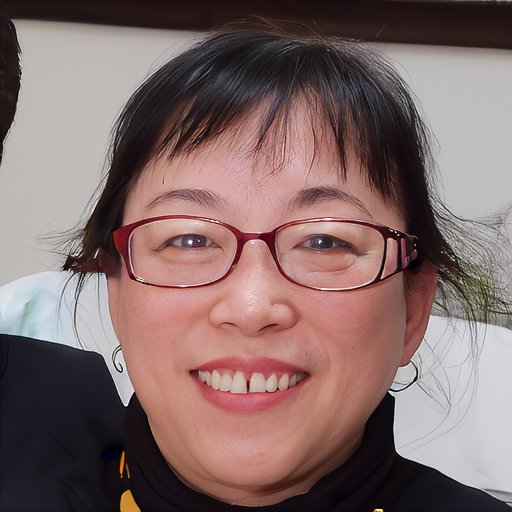}  \\
        \includegraphics[width=0.18\linewidth]{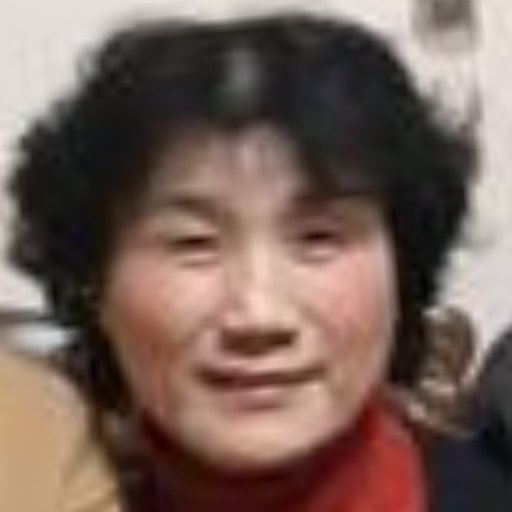} &
        \includegraphics[width=0.18\linewidth]{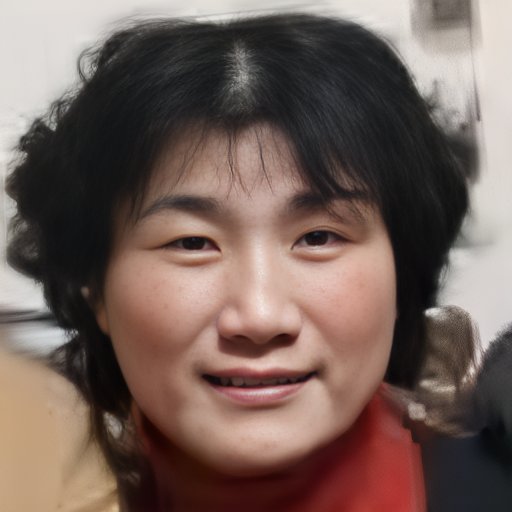}& 
        \includegraphics[width=0.18\linewidth]{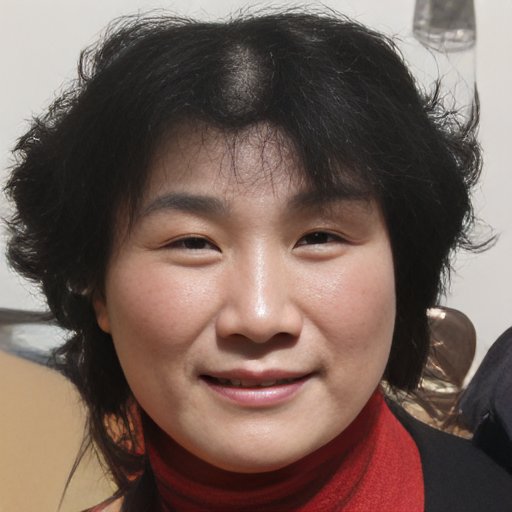}& 
        \includegraphics[width=0.18\linewidth]{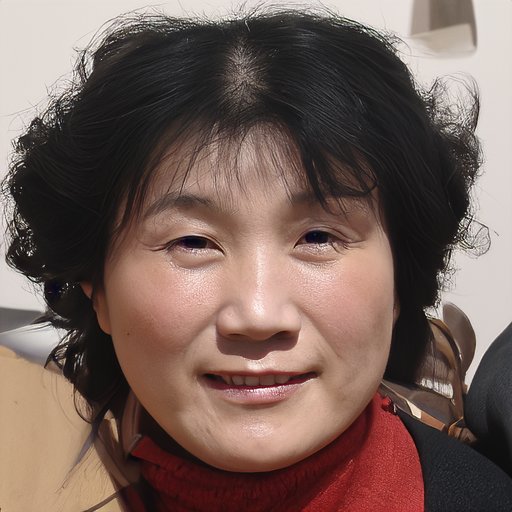}  \\ 
        \includegraphics[width=0.18\linewidth]{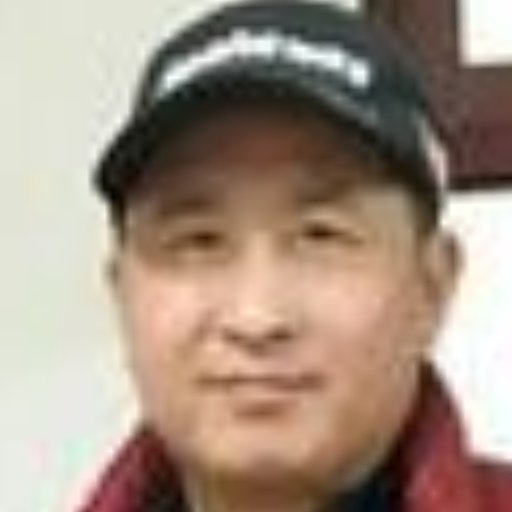} &
        \includegraphics[width=0.18\linewidth]{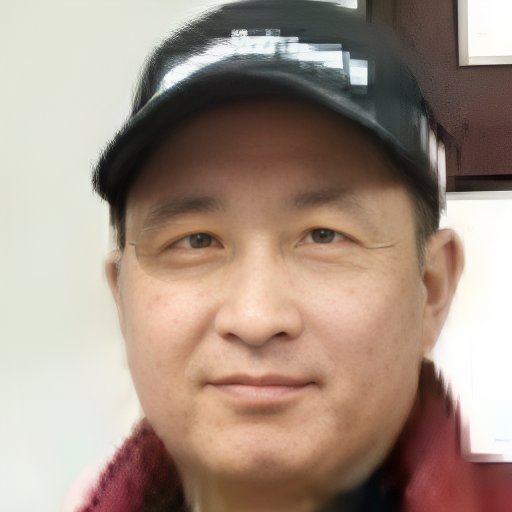}&
        \includegraphics[width=0.18\linewidth]{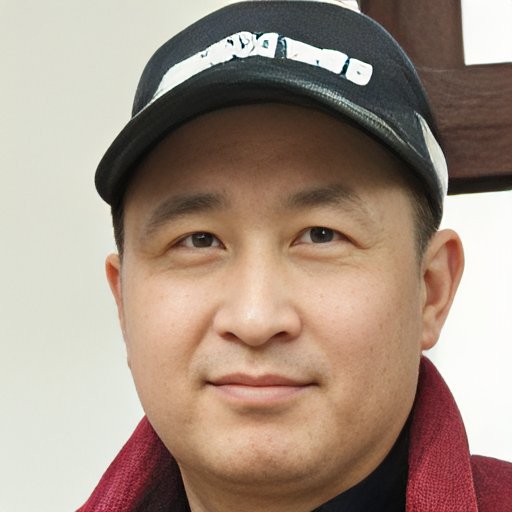} &
        \includegraphics[width=0.18\linewidth]{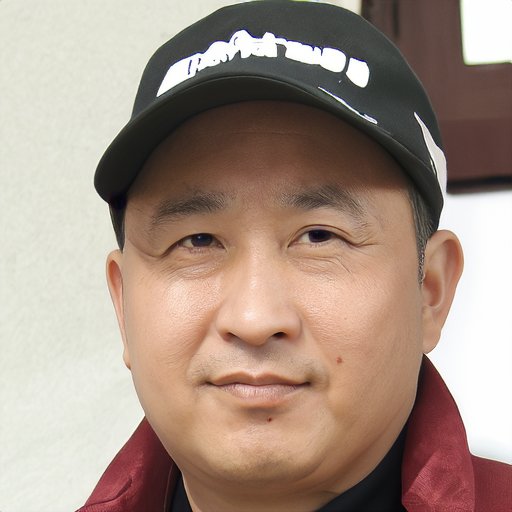} \\
        \includegraphics[width=0.18\linewidth]{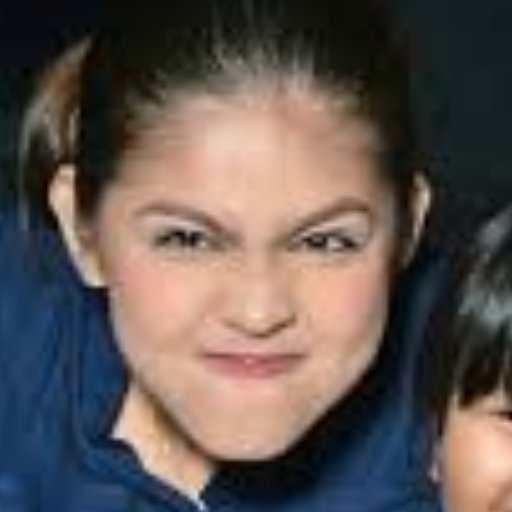} &
        \includegraphics[width=0.18\linewidth]{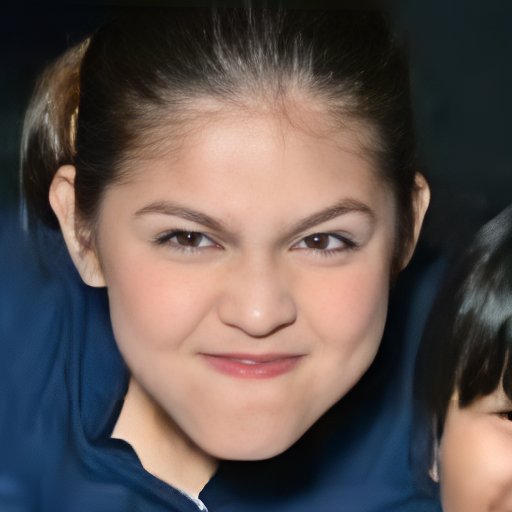}&
        \includegraphics[width=0.18\linewidth]{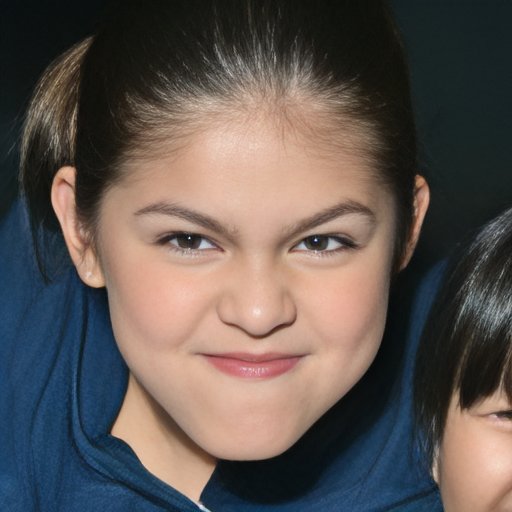} &
        \includegraphics[width=0.18\linewidth]{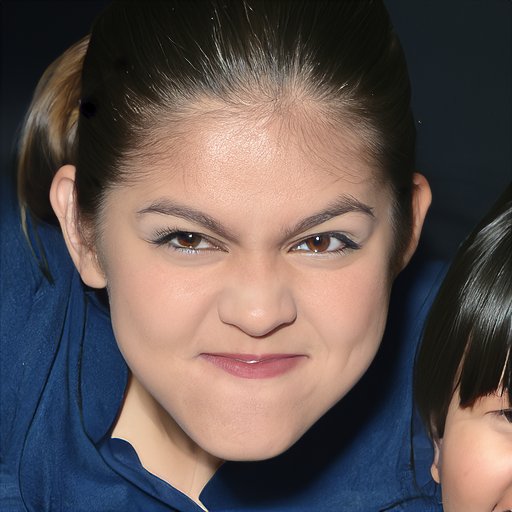} \\      
        \includegraphics[width=0.18\linewidth]{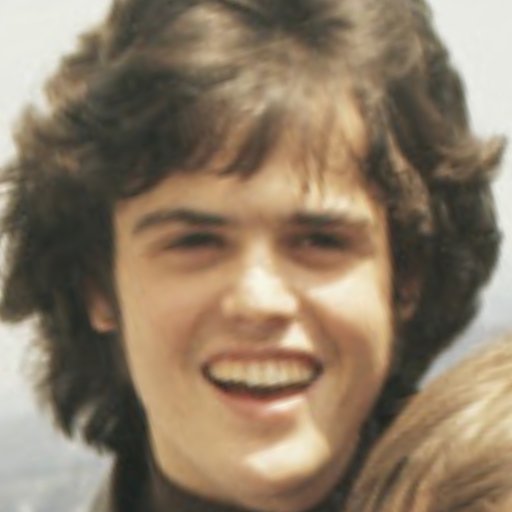} &
        \includegraphics[width=0.18\linewidth]{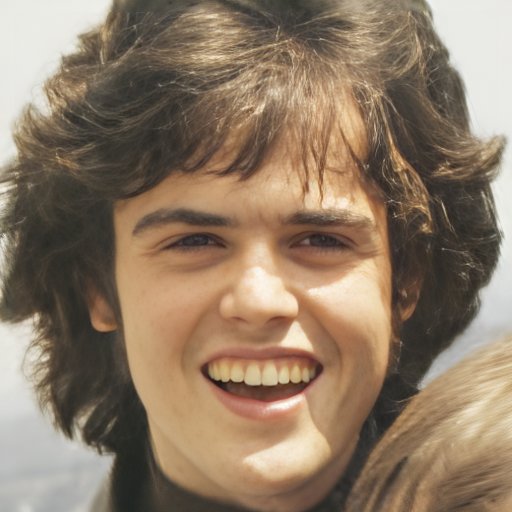}&
        \includegraphics[width=0.18\linewidth]{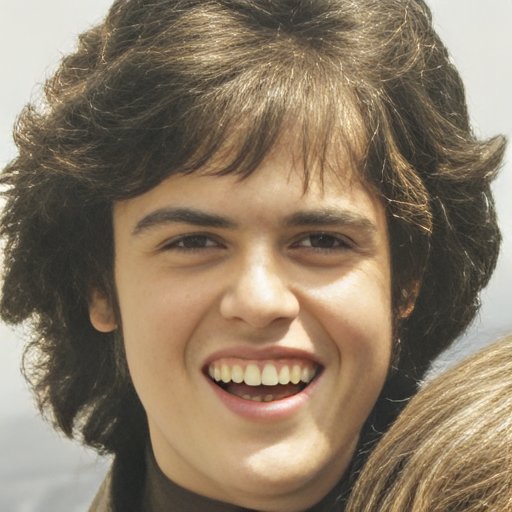} &
        \includegraphics[width=0.18\linewidth]{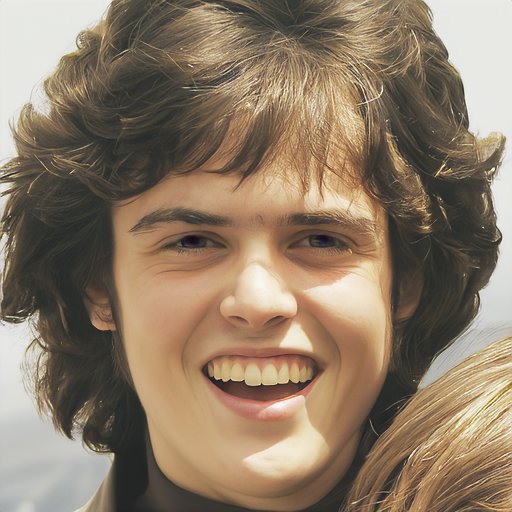} \\          
    \bottomrule
    \end{tabular}
    }
    \caption{Qualitative comparison with state-of-the-art restoration models on WebPhoto test set.}
    \label{fig:webphoto_app}
\end{figure*}

We provide more restored examples on both original CelebA-HQ (Figure~\ref{fig:celebehq_restore_app}) and synthetically degraded CelebA-HQ test datasets (Figure~\ref{fig:celebehq_deg_restore_app}), and also two real-world datasets WebPhoto (407 internet images) and LFW datasets (1711 images)~\cite{wang2021towards}, as shown in Figure~\ref{fig:webphoto_app} and Figure~\ref{fig:lfw_app}. Consistently, the proposed method can generate much more realistic, natural and faithful faces.

\end{document}